\documentclass[11pt]{article}

\usepackage[]{acl}

\usepackage{times}
\usepackage{latexsym}
\usepackage{subcaption}
\usepackage{makecell}
\usepackage{cleveref}

\usepackage[T1]{fontenc}

\usepackage[utf8]{inputenc}

\usepackage{inconsolata}

\usepackage{svg}
\usepackage{subcaption}
\usepackage{xspace}
\usepackage{tabularx}
\usepackage{euflag}
\usepackage{arydshln}

\usepackage{CJKutf8}

\usepackage{multirow}

\usepackage{booktabs}

\newcommand{\dataname}[1]{\texttt{\textbf{MAPS}}\xspace}
\newcommand{\datafullname}[1]{\textbf{M}ulticultur\textbf{A}l \textbf{P}roverbs and \textbf{S}ayings\xspace}

\newcommand{\ex}[1]{\textit{#1}\xspace}
\newcommand{\gl}[1]{``#1''\xspace}

\usepackage{mdframed}
\usepackage{lipsum}

\newcolumntype{Y}{>{\centering\arraybackslash}X}

\usepackage{xcolor, colortbl}

\definecolor{nice-red}{HTML}{E41A1C}
\definecolor{nice-orange}{HTML}{FF7F50} 
\definecolor{nice-green}{HTML}{40B5AD}
\definecolor{nice-blue}{HTML}{6495ED}
\definecolor{nice-purple}{HTML}{B990C9}
\definecolor{gray-blue}{HTML}{92c5de}
\definecolor{gray-red}{HTML}{edb7bd}

\definecolor{dgray}{HTML}{5D6D7E}
\definecolor{dblue}{HTML}{2980B9}
\definecolor{gray}{HTML}{76818c}

\definecolor{template}{HTML}{F0F0F0}
\DeclareMathSymbol{\mlq}{\mathord}{operators}{``}
\DeclareMathSymbol{\mrq}{\mathord}{operators}{`'}

\title{Are Multilingual LLMs Culturally-Diverse Reasoners? An Investigation into Multicultural Proverbs and Sayings}

\author{\bf Chen Cecilia Liu$^1$ \quad Fajri Koto$^{2}$\quad  Timothy
  Baldwin$^{2}$ \quad {\bf Iryna Gurevych$^{1,2}$ } \\
$^1$Ubiquitous Knowledge Processing Lab\\ Department of Computer Science and Hessian Center for AI (hessian.AI) \\ Technical University of Darmstadt \\
  $^2$Natural Language Processing Department, MBZUAI \\ 
{\url{www.ukp.tu-darmstadt.de}} \\
}

\begin{document}

\maketitle
\begin{abstract}
Large language models (LLMs) are highly adept at question answering and reasoning tasks, but when reasoning in a situational context, human expectations vary depending on the relevant cultural common ground. As languages are associated with diverse cultures, LLMs should also be culturally-diverse reasoners. In this paper, we study the ability of a wide range of state-of-the-art multilingual LLMs (mLLMs) to reason with proverbs and sayings in a conversational context. Our experiments reveal that: (1) mLLMs ``know'' limited proverbs and memorizing proverbs does not mean understanding them within a conversational context; (2) mLLMs struggle to reason with figurative proverbs and sayings, and when asked to select the wrong answer (instead of asking it to select the correct answer); and (3) there is a ``culture gap'' in mLLMs when reasoning about proverbs and sayings translated from other languages. We construct and release our evaluation dataset \dataname{} (\datafullname{}) for proverb understanding with conversational context for six different languages, available at \url{https://github.com/UKPLab/maps}.
\end{abstract}

\section{Introduction}

Large language models (LLMs) have achieved impressive results on question answering and reasoning tasks~\cite[inter alia]{radford2019language, brownicl, instruct}. However, when reasoning in situational context, human expectations may vary cross-culturally~\cite[i.e., pragmatic failure, the inability to understand `what is meant by what is said']{thomas1983cross} and depend on the knowledge of the relevant cultural common ground (i.e., the shared knowledge based on which people within a culture reason and communicate, including concepts, common sense, etc.~\citealp{hershcovich-etal-2022-challenges}). Understanding of such common ground in a cross-lingual setting is specifically understudied in NLP~\cite{hershcovich-etal-2022-challenges} and neglected in existing LLM literature. As languages and cultures are intertwined~\cite{kramsch2014language,hovy-yang-2021-importance}, it is crucial for models that serve all communities to be able to reason and communicate in a relevant way.

\begin{figure}[t!]
    \centering
    \includegraphics[width=0.42\textwidth]{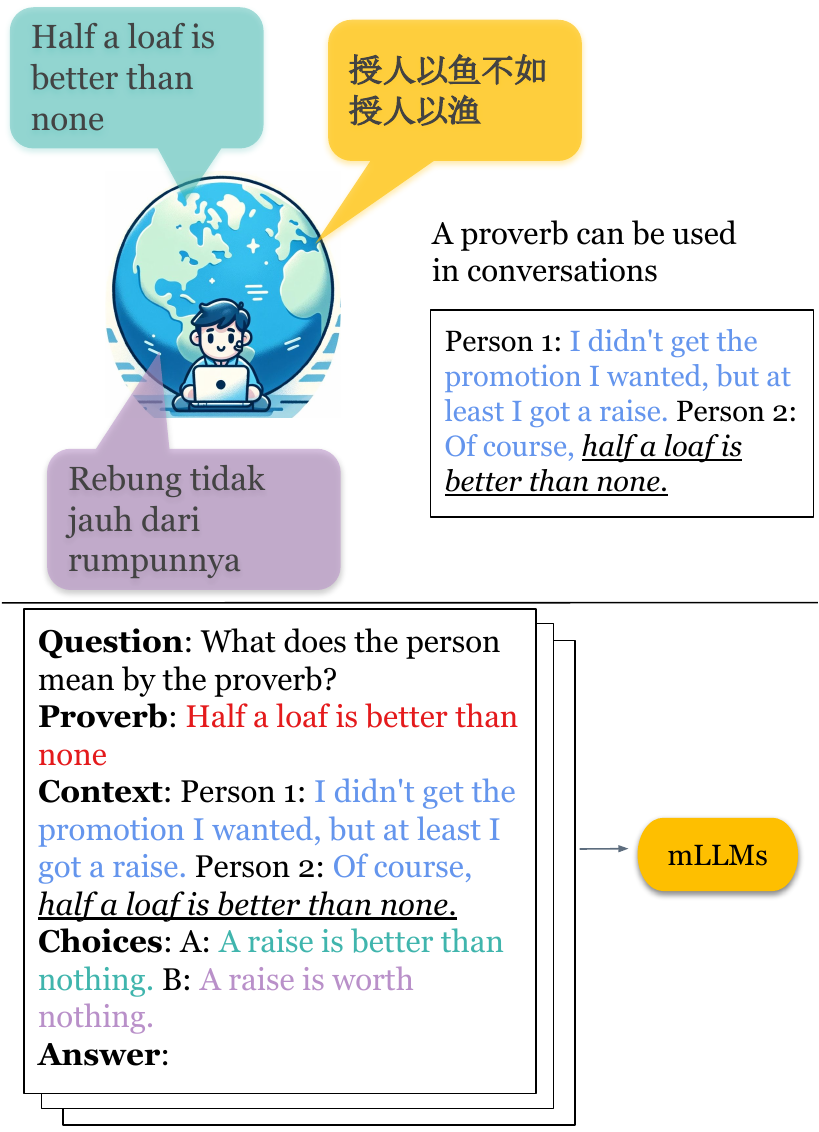}
    \caption{Proverbs are fixed expressions used by different cultures. We collect proverbs from six languages (top) and their usage within conversational contexts. We evaluate mLLMs with a binary-choice inference task in the conversational context that contains proverbs (bottom).}
    \label{fig:main}
\end{figure}
\vspace{-2.5pt}

For these reasons, we focus on studying (pragmatic) reasoning conditioned on the cultural common ground of multilingual LLMs. Several questions arise: (1) Do mLLMs embed knowledge of cultural common ground, and does this knowledge affect their reasoning performance? (2) Can mLLMs reason in contexts that require an understanding of cultural common ground? and (3) Can mLLMs reason cross-culturally (i.e., about another culture's common ground, after translating into the same language) and are there gaps in the cultural knowledge (a ``culture gap'')?\footnote{Reasoning with cultural common ground may be independent of language. For example, communications among different cultural groups within a multi-cultural country, or communication between L1/L2 speakers of a language where the L2 speaker has acquired the grammatical competence but not the cultural or pragmatic competence.}

To answer the above questions, we need to assess mLLMs using fixed, culturally-diverse expressions in multiple languages, that are also used flexibly in situational contexts. Fixed expressions are particularly important for evaluating the memorization of cultural common ground knowledge of LLMs. However, prior work focusing on multicultural concepts such as MaRVL~\cite[which is in multimodal]{marvl} or MABL~\cite{kabra-etal-2023-multi} do not contain fixed expressions. 

\begin{CJK*}{UTF8}{gbsn}

Proverbs and sayings (such as the ones illustrated in Figure~\ref{fig:main}) are fixed expressions that convey traditional wisdom, sometimes viewed as a form of folk literature and grounded in living experience and social-cultural context~\cite{white1987proverbs,mieder2004proverbs,honeck2013proverb}. While different proverbs may emerge for different cultures, the underlying meaning of proverbs usually expresses universal human experiences. Yet, their literal expression and interpretation can vary from culture to culture~\cite{honeck2013proverb}. 

For example, the English proverb \ex{{The apple doesn't fall far from the tree}} --- means a child grows up to resemble his/her parents. While a plain version \ex{like father like son} exists in many cultures, this proverb has a similar variant \ex{{Rebung tidak jauh dari rumpunnya}} \gl{Bamboo shoots are not far from the clump} in Indonesian, and \ex{{龙生龙，凤生凤，老鼠的儿子会打洞}} \gl{the dragon begets the dragon, the phoenix begets the phoenix, the son of a rat can make a hole} in Chinese. Of course, not all proverbs have parallels in different languages, as they are often culturally dependent. 
\end{CJK*}

Furthermore, proverbs are used in writing or conversational settings to offer advice, make arguments, or console others. A proverb's interpretation depends on the context~\cite{mieder2004proverbs} it is used in, and is often figurative, where the interpreted meaning does not entail the literal meaning. This makes them the ideal devices for studying the ability of LLMs to reason in situational contexts.

Hence, in this paper, we propose to use proverbs and sayings as one particular proxy for cultural common ground. In particular, we study: (1) Do mLLMs recognize proverbs, and how well do they memorize them? (2) Can mLLMs choose the correct interpretation of a proverb given a situational context? and (3) Can mLLMs reason cross-culturally, and are there culture gaps in the interpretation of proverbs across cultures?

We first present the \dataname{} (\datafullname{}) dataset, which consists of a collection of proverbs and sayings, an inference task for interpreting the meaning of proverbs in situational contexts (i.e., conversations), and binary labels indicating if the proverb is figurative. The dataset covers six languages with geographical diversity: English, German, Russian, Bengali, Mandarin Chinese, and Indonesian. 

We design a suite of experiments with \dataname{} for a wide range of \textit{open source} state-of-the-art mLLMs. We find that mLLMs do possess knowledge of proverbs and sayings to varying degrees (significantly biased toward English and Chinese), and the amount of knowledge scales with model size. Through our inference task, we also find that the memorization of proverbs does not indicate better reasoning ability with proverbs, and figurative proverbs are more difficult for mLLMs to reason about in many languages. On the ability of mLLMs to reason cross-culturally with cultural common ground, we find that significant culture gaps exist when reasoning with translations. Our results indicate that despite the apparent multilingual reasoning abilities of mLLMs, further research to improve the cultural diversity (in terms of cultural common ground) of mLLMs is needed. 

To summarize, our contributions are: \textbf{(1)} we provide an analysis of the ability of a wide range of state-of-the-art open-source mLLMs to reason with cultural common ground, through the lens of proverbs and sayings; \textbf{(2)} We disentangle the effects of memorization versus reasoning with proverbs and sayings, and reveal culture gaps in mLLMs; and \textbf{(3)} We construct a multicultural dataset of proverbs and sayings for six different languages with multiple levels of annotations. 

\section{Related Work}
Prior work has evaluated the ability of  LLMs to reasoning  abstractly~\cite[recognize proverbs from short stories]{ghosh-srivastava-2022-epic} or inference based on cultural norms~\cite{huang-yang-2023-culturally} in English and assessed the models' ability for matching proverbs across three languages~\cite[with a small evaluation set]{bigbench}. To the best of our knowledge, \dataname{} is the largest multilingual dataset that focuses on proverbs and sayings, with conversational contexts and an inference task. 

MABL~\cite{kabra-etal-2023-multi} is a task similar to ours but focuses on the multicultural understanding of novel metaphors and cross-lingual transfer. It is less suitable for studying memorization versus reasoning and does not study reasoning within a conversational context. \citet{lauraruis} and \citet{hu-etal-2023-fine} use conversational context to study pragmatic reasoning in English LLMs and the identification of parallels between humans and models, respectively. Concurrently, \citet{huang-yang-2023-culturally} proposed a culturally-aware natural language inference task based on cultural norms. However, they provide limited insights beyond English. While we also use conversational context, we focus on cultural common ground and multilingual aspects of mLLMs (with a larger dataset). Other work on understanding the memory-retrieval mechanism in LLMs with English idioms~\cite{haviv-etal-2023-understanding}, cultural knowledge~\cite{SeaEval2023, indommlu, cmmlu} or cultural value and bias~\cite[inter alia]{arora-etal-2023-probing, haemmerl-etal-2023-speaking, cao-etal-2023-assessing}. Furthermore, we acknowledge existing work intended to study the formal and other types of reasoning in LLMs (such as the ones mentioned in \citealp{huang-chang-2023-towards}), which are different in their goals from ours.

\section{MAPS --- \datafullname{}}\label{sec:data}

To help investigate our proposed research questions, we first present \dataname{} --- a dataset of proverbs across six geographically and topologically diverse languages. \dataname{} consists of: (1) proverbs and sayings; (2) conversational usages as context; (3) interpretations of proverbs (one correct, one wrong); and (4) labelling of whether the usage of the proverb is figurative or not (see Table~\ref{tab:examples} for data examples, and Figure~\ref{fig:annotation} in Appendix~\ref{app:examples} for an illustration of the annotation process).

\subsection{Dataset Creation}

\paragraph{Language Choices.} We chose six languages for this dataset: English, German, Russian, Bengali, Mandarin Chinese, and Indonesian. Several factors were considered when choosing the languages, including geographical diversity such as Eastern vs.\ Western (to increase the potential concept diversity), topological diversity, and resource availability (high-resource vs.\ lower-resource).

\paragraph{Proverbs and Sayings.} We collect all proverbs and sayings (along with explanations) from Wikiquote\footnote{\url{https://en.wikiquote.org/}} and Wiktionary.\footnote{\url{https://www.wiktionary.org/}} Bengali has a significantly higher volume of proverbs compared to other languages, and thus we perform random sub-sampling of the proverbs for annotation to keep the final data roughly balanced between languages.

\paragraph{Conversational Context.} While proverbs and sayings are self-contained, they are typically used in conversations and writing. To investigate the ability of mLLMs to reason with proverbs, next, we created short conversations that use proverbs (i.e., the conversational context for the inference task). 

To aid the data creation process, we use a model-in-the-loop approach, inspired by recent work~\cite{chakrabarty-etal-2022-flute, afraid}. We first use GPT3.5 (gpt-3.5-turbo-0301; a sibling model of~\citealp{instruct}) by prompting it with fixed templates to generate the seed conversational context (see Appendix~\ref{app:template} for the model templates).\footnote{The conversational contexts are in each perspective language, except for Russian and Bengali where the contexts are in English due to quality issues. For Russian and Bengali, the contexts are written in English first, then machine-translated and fixed by native speakers for two rounds.} Next, we ask two or more native speakers (experts or crowd, with at least one expert per language) to either accept the model-created conversation or write a new conversation if the usage of the proverb is flawed.

In the final dataset, the conversational contexts for English, Chinese, Russian, and Bengali were completely rewritten,\footnote{The model has significant trouble in creating relevant context when the proverb is figurative. Anecdotally, human annotators found that the machine-generated context is helpful as a `prompt', which helped to speed up the rewrites.} whereas for Indonesian and German, 22\% and 20.5\% of the original model-generated contexts were retained (the difference is probably due to variations in individual annotator preferences). 

\paragraph{Interpretation of Proverbs in Context.} We formulate this part as an inference task (following \citealp{figqa}). We ask annotators to create one correct answer and one wrong answer to the following question based on the conversational context: 

\textcolor{nice-blue}{\textit{What does the person mean by \{proverb\}?}}

Additionally, we also label the proverb if the interpretation is figurative (i.e., the interpreted meaning of the proverb is different from the expressed literal meaning).\footnote{\gl{An apple a day keeps the doctor away} is a literal proverb that advocates for apple consumption. \gl{The apple doesn't fall far from the tree} is a figurative proverb where the literal meaning is about apples and a natural phenomenon, whereas the actual meaning of the proverb is about a child growing up to resemble his/her parents.}

\paragraph{Quality Control.}
Finally, we sampled 100 conversational contexts with their answers from each language. Then, we asked a separate set of native speakers to assess the data quality for: (1) correct usage of the proverb (i.e., the context is correct); and (2) correct answers for interpreting the meaning. Sometimes, it is possible to have more than one interpretation of a proverb given the context. We asked the native speakers to score the answers as correct as long as the answers aligned with one possible interpretation and revise the options. 

The final dataset consists of 2313 proverbs with conversational context. The statistics for each language are in Table~\ref{tab:data_char} (with additional data statistics in Table~\ref{tab:data_stats} in Appendix~\ref{app:dataset}). We further split the data for each language into a test set and a few-shot train-dev set (30 randomly selected examples each). Table~\ref{tab:examples} shows examples from our dataset. 

\begin{table}[]
    \centering
    \resizebox{0.88\linewidth}{!}{%
    \begin{tabular}{lccc}
    \toprule
         \textbf{Language} & \textbf{Code} & \textbf{\#Data (Test Size)} & \textbf{Class} \\
         \midrule
         English & En & 424 (394) & 5\\
         Chinese & Zh & 364 (334) & 5\\
         German  & De & 364 (334) & 5\\
         Russian & Ru & 420 (390) & 4\\
         Bengali & Bn & 370 (340) & 3\\
         Indonesian & Id & 371 (341) & 3 \\
    \bottomrule
    \end{tabular}
    }
    \caption{Dataset statistics. ``Class'' = language class according to~\citet{joshi-etal-2020-state}, where 5 means the language is resource-rich.}
    \label{tab:data_char}
\end{table}

\subsection{Analysis of \dataname{}}
\label{subsec:dataset_char}

Proverbs and sayings are cultural artifacts and reflect embodied experiences, which contain diverse concepts often grounded in the real world.
For instance, dairy product concepts (milk, cheese, yogurt, etc.) exist in different languages but not in Chinese proverbs, whereas concepts that are symbolically meaningful in Chinese culture like dragons or phoenixes exist in the dataset. To illustrate this, we select interesting food items and animals from the final dataset (details in Table~\ref{tab:data_term}, Appendix~\ref{app:data_analysis}). Furthermore, we categorized the concepts in 100 sampled figurative proverbs in English, Chinese, and Indonesian (for details, see Appendix~\ref{app:qual_analysis}, Figure~\ref{fig:pie_fig}). We observe that Indonesian has a lot more proverbs that use animals and are about nature than English.  

\begin{figure}
    \centering
    \includegraphics[width=0.43\textwidth]{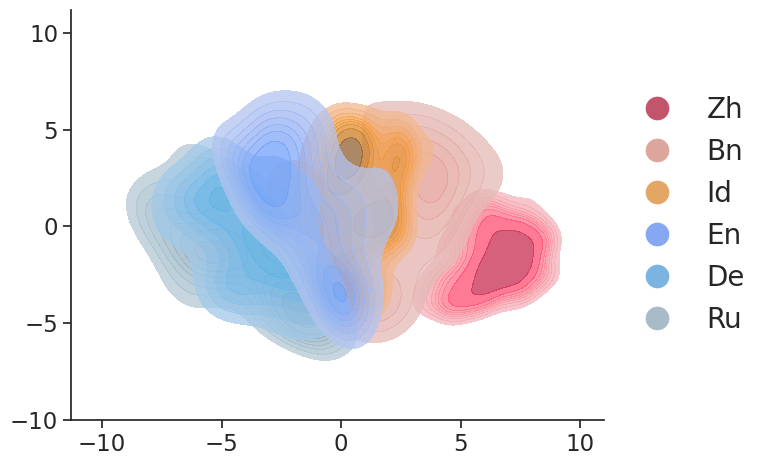}
    \caption{Visualizing proverb embeddings using kernel density estimation (KDE).}
    \label{fig:data_analysis}
\end{figure}

We further encode the proverbs (without contexts) using multilingual sentence embeddings~\cite[LaBSE]{labse} and plot the embeddings with Kernel Density Estimate (KDE) (after dimensionality reduction to two components using tSNE;~\citealp{tsne}) to show the distinctiveness and connections between proverbs across different languages and cultures in Figure~\ref{fig:data_analysis}, which further illustrates that proverbs and sayings are inherently culturally-diverse. To verify this is not due to language difference, we provide additional analysis and discussion in Appendix~\ref{app:kde} to isolate the language effect.

In Figure~\ref{fig:data_analysis}, the embedding distributions are interestingly ordered from the West to the East. Indonesian proverbs partially overlap with English, probably due to the use of the Latin script and influences of foreign languages due to historical context. Chinese and Bengali proverbs are relatively distinct from the Western languages. 

\begin{CJK*}{UTF8}{gbsn}

\begin{table*}[t!]
    \centering
    \resizebox{0.92\textwidth}{!}{%
    \begin{tabular}{llll}
        \toprule
        \textbf{Lang} 
        & \textbf{\textcolor{nice-purple}{\textbf{Proverb}}}
        & \textbf{\textcolor{nice-green}{\textbf{Context}}}
        & \textbf{\textcolor{nice-orange}{\textbf{Choices}}  \& \textcolor{nice-blue}{\textbf{Answer}} }

        \\\midrule

        \makecell*[l]{Zh\\}
        &
        \makecell*[{{p{2.cm}}}]{
        授人以鱼不如授人以渔 \\ (figurative)}
        &
        \makecell*[{{p{8.5cm}}}]{
        A: 你可以帮我做这个项目吗？ B: 当然可以，但是我觉得“授人以鱼不如授人以渔”。\\
        \texttt{(A: Can you help me with this project? B: Of course, but I think "it is better to teach a man fishing than to give him fish".)}\\
        }
        &
        \makecell*[{{p{8cm}}}]{
        A: B想帮A做项目而不是教A做项目。 \\
        \texttt{(B wants to help A with the project instead of teaching A to do the project.)}\\
        B: B想教A做项目而不是帮A做项目。 \\
        \texttt{(B wants to teach A to do the project instead of helping A to do the project.)}\\
        \textcolor{nice-blue}{\texttt{\textbf{Answer: B}}}
        } \\\midrule
  
        \makecell*[l]{Id\\}
        &
        \makecell*[{{p{2.cm}}}]{
        Nasi sudah menjadi bubur \\(figurative)}
        &
        \makecell*[{{p{8.5cm}}}]{
        Orang 1: Bagaimana reaksi bos-mu setelah kamu mengakui kesalahanmu? Orang 2: Kurang baik. Saya sudah mencoba menjelaskan alasan saya berbuat begitu, tetapi saya tetap diberi sangsi. Nasi sudah menjadi bubur. \\
        \texttt{(Person 1: How did your boss react after you admitted your mistake?
                Person 2: Not well. I've tried to explain why I did this, but I'm still being penalized. The rice has become porridge.)}}
        &
        \makecell*[{{p{8cm}}}]{
        A: Orang 2 tidak dapat melakukan apapun untuk mengubah reaksi bos.\\
        \texttt{(Person 2 can do nothing to change the boss's reaction.)}\\
        B: Orang 2 masih bisa mengubah reaksi atasan.\\
        \texttt{(Person 2 can still change the boss's reaction.)}\\
        \textcolor{nice-blue}{\texttt{\textbf{Answer: A}}}
        } \\
        \bottomrule
    \end{tabular}}
    \caption{Examples from selected languages (see Table~\ref{tab:all_examples}, Appendix~\ref{app:examples} for examples in all languages).}
    \label{tab:examples}
\end{table*}

\end{CJK*}

\section{Experimental Setup}
\label{sec:exp}
We perform zero-shot evaluations and keep all prompt templates in English (on the test set), as previous studies show better performance with English prompts on mLLMs~\cite{xglm, bloom, muennighoff-etal-2023-crosslingual}.\footnote{For completeness, we also provide additional baselines using \dataname{} for cross-lingual transfer and few-shot evaluation in Appendix~\ref{app:xling_trans} and Appendix~\ref{app:nshot}.}

\paragraph{Models.} 
We experiment with the following open source state-of-the-art multilingual models: (1) masked LMs (MLM): XLM-R (355m, 3.5B, \citealp{xlmr}); (2) encoder--decoder LMs: mT0 (580M, 3.7B, 13B, multitask and instruction tuned, \citealp{muennighoff-etal-2023-crosslingual}); and (3) Causal LMs: BLOOMZ (560M, 3B, 7.1B, \citealp{muennighoff-etal-2023-crosslingual}), and XGLM (564M, 2.9B, 7.5B, \citealp{xglm}). Most of the models cover all 6 languages in \dataname{} except BLOOMZ, which is derived from BLOOM~\cite{bloom} and does not cover Russian or German. In addition, despite being primarily an English model, LLaMA-2~\cite[Causal LM]{llama2} has some multilingual capabilities. As a result, we incorporate three LLaMA-2 models (7B, 13B, 70B) in our study.\footnote{While larger models exist, we chose these models due to computational constraints. We can already see differences in performance at these model sizes and we include additional results for Vicuna-v1.5~\cite{vicuna1p5} and Aya-101~\cite{ayamodel} in the Appendix~\ref{app:add_models}.} 

\paragraph{Memorization Evaluation.} 
Since proverbs are \textit{fixed} expressions, successfully completing a proverb with greedy decoding likely means that the model has seen the proverb during pretraining, similar to prior work on detecting memorization or data contamination in LLMs~\cite{magar-schwartz-2022-data, haviv-etal-2023-understanding, carlini2023quantifying, DBLP:conf/uss/CarliniTWJHLRBS21}.\footnote{See Appendix~\ref{app:add_mem_disc} for more discussion.} Hence, following a similar setup to previous work~\cite{magar-schwartz-2022-data, haviv-etal-2023-understanding, carlini2023quantifying, DBLP:conf/uss/CarliniTWJHLRBS21}, we mask out the last word of each proverb and prompt the mLLMs to complete the proverb with templates in Table~\ref{tab:prompt_temp_mem}, Appendix~\ref{app:template}.

For the memorization task, let $t_i \in T$ be a prompt template, and let $q_j$ be a proverb with \textit{n} words where $q_j \triangleq \{w_1, w_2 \cdots w_n\}$. We remove the last word $w_n$ for non-MLM models, if the LM generates (greedily) a string that starts with the missing token, or the entire proverb is a sub-string of the generated string, then we count the model as having memorized the proverb. For the MLM model, we mask out the last word with `<mask>' and do predictions (i.e., $w = \arg\max_{w_n \in V} P(w_n | T_i; \hat{q}_j)$, where $\hat{q}_j$ is a proverb with mask token, and $V$ is the vocabulary). 

As the zero-shot prompting results are highly sensitive to the input patterns, we create 5 different prompt patterns (Table~\ref{tab:prompt_temp_mem}, Appendix~\ref{app:template}), and take the union of memorized examples among 5 patterns as the memorization accuracy. 

\paragraph{Reasoning Evaluation.} For the inference task, we compute the correct answer by comparing the logits of the two answer candidates (`A' or `B') as in \citet{xglm}. In particular, we use the prompt template $t^{r}$ for this task (as in Table~\ref{tab:prompt_temp}, Appendix~\ref{app:template}) and compute $ P(t^{r};q_i;\mlq A\mrq)$ and $P(t^{r};q_i;\mlq B\mrq)$ and pick the larger one as the correct answer. For the MLM model, we compare the prediction logits of the answer candidates. 

\paragraph{Translations for Cross-culture Gap Evaluation.}\label{method_trans} To study gaps in cross-cultural communication, we use English and Chinese as the basis for a case study, with two types of translation data. 
\textbf{Machine Translation (MT).} We translate every Chinese proverb, context and answer into English using Google Translate (Zh--En). 
\textbf{Human-Adapted Translation (HT).} We perform adaptations to the machine-translated context: (1) manually correct any mistakes in the literal translation of proverbs, fixing the grammatical errors in the contexts and answers; and (2) conduct a light adaptation of the translated data inspired by~\citet{majewska-etal-2023-cross}, by replacing names and locations in the dataset to align with the culture (e.g., XiaoMing to Michael etc.) in case LLMs are confused about whether an entity is a person or a place. This represents our best-effort adaptation to reduce the language gap.

\begin{figure*}
    \centering
    \begin{subfigure}[b]{\linewidth}
         \centering
         \includegraphics[width=0.98\linewidth]{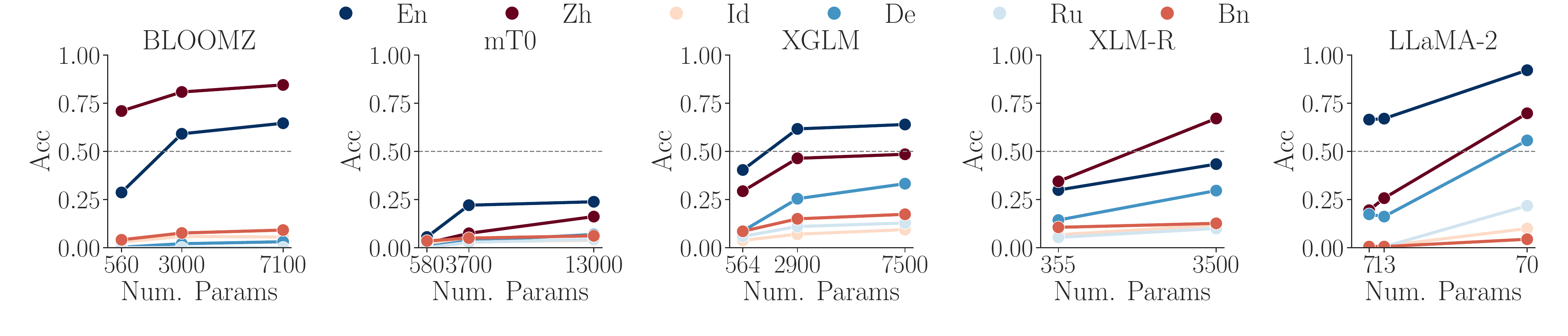}
         \caption{Memorization of proverbs in different languages.}
         \label{fig:mem1}
     \end{subfigure}
    \hfill
    \vspace{0.2cm}
    \begin{subfigure}[b]{\linewidth}
         \centering
        \includegraphics[width=0.98\linewidth]{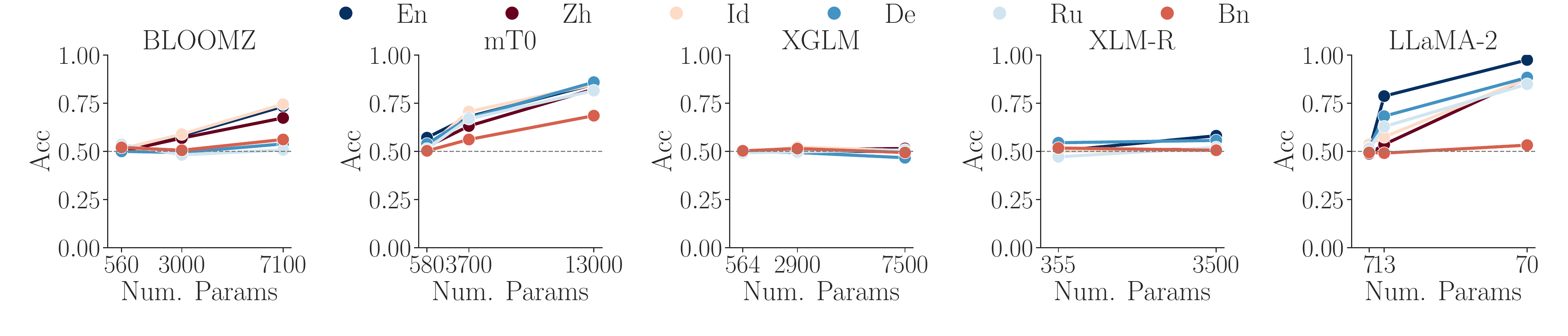}
        \caption{Zero-shot results of proverbs understanding with context.}
         \label{fig:qa_pos}
     \end{subfigure}

    \caption{Performance of mLLMs on the proposed \dataname{} dataset. The number of parameters is in billions for LLaMA-2 and in millions for all other models.}
    \label{fig:mem}
\end{figure*}

\section{Results and Discussion}

\subsection{Knowledge of Proverbs} 
\textit{--- A little knowledge is a dangerous thing.}

While it is possible that the proverbs in the training data appear alone without any contextual usage or explanation, we consider such an occurrence to be unlikely.\footnote{Webpages such as this \url{https://en.wiktionary.org/wiki/no_pain,_no_gain} exist in the training data for LLMs.} Hence, we make the assumption that memorization of the fixed expression also correlates with LLMs having embedded knowledge of the usage or meaning.

Figure~\ref{fig:mem1} shows the results of proverb memorization, which (unsurprisingly) improves with model size. While XLM-R, XLGM, and mT0 cover all of the languages in our dataset, they don't score particularly well in memorization of proverbs in a single language. All models exhibit disparities in memorization across all languages, and these disparities are particularly pronounced in the case of Indonesian, Bengali, and Russian, which are lower-resource languages.\footnote{See Appendix~\ref{app:add_mem_res} for more results.} These disparities are potentially due to data exposure, as we don't find any significant attribution, such as well-known versus less well-known, long versus short, or figurative versus non-figurative proverbs, by analyzing the memorized proverbs.

\subsection{Reasoning of Proverbs with Conversational Context}
\textit{--- All that glitters is not gold.}

While many models embed knowledge about proverbs, it is unclear if memorization translates to better reasoning with proverbs given the context. Next, we assess the models using our inference task. 

\paragraph{Memorization does not indicate the ability to reason with proverbs.} We prompt models with the pattern in Table~\ref{tab:prompt_temp} (Appendix~\ref{app:template}) and plot the accuracy across languages in Figure~\ref{fig:qa_pos}. In general, the bigger the model is, the better it performs on the inference task (i.e., the ability emerges with scale). 

Overall, comparing the memorization curve and reasoning curve of mT0, XGLM and XLM-R, we observe that memorization does not indicate the ability to reason with proverbs in our experiments. In fact, model architecture has little effect (as BLOOMZ and LLaMA-2 are Causal LMs, and mT0 is an encoder--decoder model). We provide additional results using different prompts in Appendix~\ref{app:add_pos} and few-shot results in Appendix~\ref{app:nshot}.

Since we know which proverbs are memorized from the previous experiments, we further break down the results into memorized vs.\ not memorized proverbs for the 3 best-performing models excluding LLaMA-2 70B (as it already achieved good results in Figure~\ref{fig:mem}, which offers limited insights) in English and Chinese in Table~\ref{tab:mem_vs_reason}, Appendix~\ref{app:mem}. The benefit of memorization is evident in English and shows inconsistency in Chinese (which aligns with observations for other languages in Figure~\ref{fig:qa_pos}).

One possible explanation for the task not being heavily dependent on memorization is that contextual information aids inference, and the model may also implicitly learn other culturally-relevant information from the training data during pretraining. Consequently, this suggests that LLMs may prioritize contextual information over memory retrieval when both are available. However, such a hypothesis requires further research, which we leave to future work.

\paragraph{Figurative proverbs are difficult to understand in general.} Many proverbs are figurative, hence, we further divide the results of the model based on this property (described in \S\ref{sec:data}). Looking at Table~\ref{tab:figurative}, we can see that, across English, German, and Russian, all models perform worse on the inference task when the interpretation is figurative. Interestingly, the opposite pattern is consistently observed for Chinese. Larger models appear to understand Indonesian and Bengali figurative proverbs better. One conjecture is that while abstract reasoning (the kind required for understanding figurative proverbs) can rely on memorization, less memorization may lead to better abstract reasoning in LLMs.

\begin{table*}[]
    \centering
        \resizebox{0.91\linewidth}{!}{

    \begin{tabular}{l|cccccc}
    \toprule
    \textbf{Model} & \multicolumn{6}{c}{Non-Figurative / Figurative} \\
            & En & Zh & Id & De & Ru & Bn\\
    \midrule
        BLOOMZ 3B   & 58.76/57.60 &  53.12/61.97 & 53.33/60.52 & \textcolor{gray}{51.66/47.54} &\textcolor{gray}{52.43/45.13} &  55.88/49.26  \\  
        BLOOMZ 7.1B   & 79.66/68.20 & 66.66/68.30 & 72.00/75.18 & \textcolor{gray}{54.30/53.55} & \textcolor{gray}{52.43/49.55} & 67.64/53.30  \\
        mT0-XL (3.7B)   & 75.14/62.21 & 62.50/64.08 & 74.67/69.54 & 74.17/61.74 & 73.78/61.94 & 69.12/52.94 \\
        mT0-XXL (13B)   & 87.01/82.95 & 81.77/83.09 & 84.00/84.96 & 88.74/83.61 & 87.80/76.99 & 63.23/69.85 \\
        LLaMA-2 13B   & 81.36/76.50 & \textcolor{gray}{53.12/54.23} & \textcolor{gray}{54.66/58.27} & \textcolor{gray}{72.19/65.03} & \textcolor{gray}{67.07/59.73} & \textcolor{gray}{47.05/49.63} \\
    \bottomrule
    \end{tabular}
    }
    \caption{Zero-shot accuracy of non-figurative and figurative proverbs (Non-Fig./Fig.). The gray colour results indicate that the language is not officially supported by the model.}
    \label{tab:figurative}
\end{table*}

\begin{figure*}
    \centering
    \includegraphics[width=0.98\linewidth]{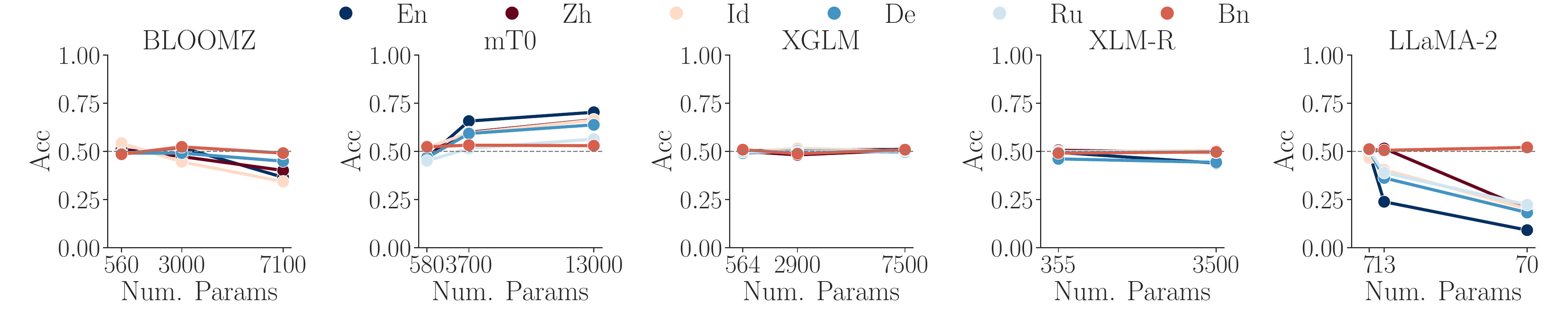}
    \caption{Performance of mLLMs on the proposed \dataname{} - Inference task when asking the `negative' question. The number of parameters is in billions for LLaMA-2 and in millions for all other models.}
    \label{fig:qa_neg}
\end{figure*}

\paragraph{Bias towards the correct answer amplifies performance gaps across languages.} If the model genuinely understands a proverb's meaning in a situational context, it should be able to select the correct answer as well as the wrong answer when requested, especially for a task with only two choices. Prior work has shown that negation in the natural language inference task weakens model performance~\cite{truong-etal-2023-language, she-etal-2023-scone, kassner-schutze-2020-negated}. While not the primary focus of our work, this is a fundamental aspect of reasoning~\cite{blanco-moldovan-2011-semantic} and we conducted experiments to verify. Here, we aim to ask a `negative' question rather than provide negative answers. Hence, we change the question in the prompt template to \ex{What does the person \textit{not} mean by the proverb?}, while keeping everything else the same.

The results are in Figure~\ref{fig:qa_neg}. By simply asking the model to pick the wrong answer, all previously well-performing models are now performing badly, except mT0 (which may be due to the model being instruction-tuned). The `negative' question enlarged performance gaps across languages as the model size increased. Additional results on asking the model to pick the wrong answer \textit{without} using the word \textbf{\textit{not}} are in Appendix~\ref{app:negation}, where we observe consistent trends of model failures and inverse scaling in many cases. While we focus on the cultural aspect of mLLMs, these results show fundamental work is needed to improve the ability of current mLLMs to handle `negative' questions.

\subsection{Culture Gaps in mLLMs - A Case Study}\label{sec:cult_gap}
\textit{--- When in Rome, do as the Romans do.}

An ideal mLLM should perform on texts from all languages and translations in all directions equally well. However, in our experiments, the performance on English data is still stronger than in other languages for most of the models we studied. Recently, several works have shown that good performance can be achieved by translating non-English text data in languages into English~\cite[inter alia]{conneau2019cross,yang-etal-2019-paws}. 
Here, we demonstrate that when a task relies on cultural context, there are two distinct performance gaps to achieving true multilingual ability: one is the language gap (due to mistakes by the translation system, which may be fixed by a perfect translation system), and the other is the culture gap.\footnote{This can relate to cross-cultural pragmatic failure.} To demonstrate this, we use English and Chinese as the focus of a case study. 

\begin{figure}
    \centering
    \includegraphics[width=0.5\textwidth]{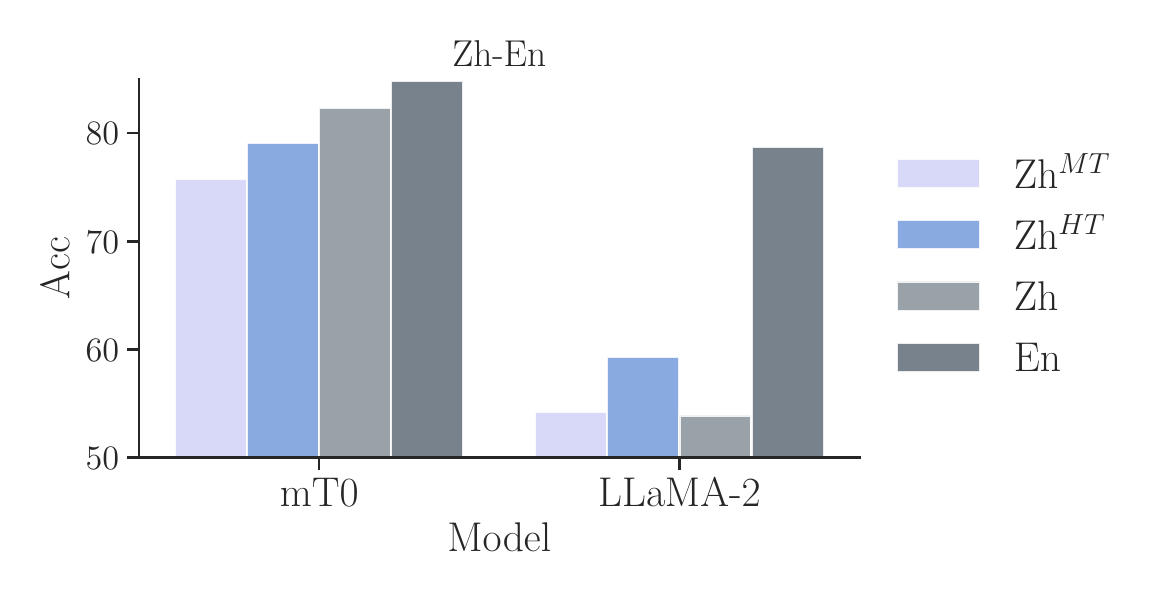}
    \caption{Performance gap between machine-translated, human-translated data and results in the original source language (Zh), and target language (En).}
    \label{fig:zh2en_cult_gap}
\end{figure}

We translate the data based on the descriptions in \Cref{method_trans}. Next, we perform the zero-shot evaluation with the best-performing multilingual models (mT0-XXL, 13B) and English model (LLaMA-2 13B) for Zh--En (in Figure~\ref{fig:zh2en_cult_gap}). In fact, both models show a performance gap in the translated data compared to the target language. Interestingly, mT0 also shows a performance degradation compared to the inference results in the original language (as LLaMA-2 is near chance level for Zh, the improvement is not surprising). In all cases, HT improves over MT, where the gain can be considered as the language gap. More interestingly, we define the gap between HT and the $\max$ of source and target language is the \textit{culture gap} in mLLMs, i.e., \textit{culture gap} = $|\textrm{Acc}^{HT} - \max (\textrm{Acc}^{Src},\,\textrm{Acc}^{Tgt})|$. The culture gap for Zh--En is 5.73 for mT0 and 19.40 for LLaMA-2.\footnote{We also perform the same experiment in the reverse direction En--Zh with mT0 (Appendix~\ref{app:trans_test}), similar results were observed. Other evaluation results on machine-translated data for other languages with LLaMA-2 are in Appendix~\ref{app:trans_test}.} In an ideal situation, these gaps should be 0, indicating that the model is culturally aware and capable of understanding a language when speakers come from diverse cultural backgrounds. 
\begin{CJK*}{UTF8}{gbsn}
By closely examining the machine-translated data, it is evident that current machine translation (MT) systems do not handle cultural context well, producing incomplete or incorrect translations of proverbs. For example, a polysemous phrase 大三 was translated to \gl{junior} (third-year university student), but in a specific proverbial context, it means someone is \gl{three years older}. Similarly, a phrase like 不如 ~is translated to \gl{not as good as} instead of \gl{it'd be better}. Our results suggest that additional research is needed to improve cultural awareness and the inclusion of cultural priors in MT models and mLLMs~\cite{culttrans, shaikh-etal-2023-modeling}.
\end{CJK*}

\section{Conclusion}
In this work, we use proverbs and sayings from different languages as an investigative tool to assess the ability of mLLMs to reason with cultural common ground. Specifically, we study various mLLMs to evaluate their ability to memorize proverbs, reason with proverbs and sayings in different situational contexts and understand cross-cultural communications using proverbs.

To aid the investigation, we developed a multicultural proverbs and sayings dataset \dataname{}. 
Our analysis shows that many models possess knowledge of proverbs and sayings, however, recognizing proverbs does not mean the model is able to reason with proverbs in contextual settings. 
Indeed, we found that mT0 shows some culturally-diverse reasoning ability, but only to a very limited extent. We also found that the ability to reason in a zero-shot manner emerges with model scale, but the ability to understand a `negative' question inversely correlates with the model scale. The disparities in culturally-diverse reasoning ability between languages grow with the model size, which raises concerns in terms of multilingual availability and points to the need for more robust mLLMs. Finally, we defined and observed several culture gaps in cross-lingual communications. We hope to explore different aspects of cultural common ground in the future and to inspire novel work that facilitates inclusive cross-cultural understanding and communication with mLLMs.

\section{Limitations}

Our work uses proverbs and sayings as a proxy for cultural common ground, and we explore mLLMs' ability to understand cultural common grounds in a limited setting. One potential limitation is we only collect one conversation per proverb or saying, and one pair of correct--wrong interpretations. Another limitation is the evaluation data is relatively small compared to many automatically generated benchmarks and may introduce lexical biases. However, these are not major concerns as: (1) we want to focus on cultural common ground, which automatically limits us to a subset of lexical items (lexical biases is an intended feature); and (2) to the best of our knowledge, this is the largest proverb dataset for reasoning in context, and there is enough signal to distinguish between the tested models and uncover insights on current mLLMs ability and limitations in understanding proverbs and sayings. We hope to explore aspects of culture beyond proverbs and sayings, and with a more diverse set of languages (such as African languages or American indigenous languages) in the future.

In this work, we evaluate general-purpose open-source mLLMs. However, a full evaluation of larger models or task-specific models may be necessary, especially when asking `negative' questions and assessing culture gaps. We focus on studying open-source LLMs in this paper for scientific reproducibility, and closed-source LLM evaluations are beyond our scope. As our dataset is publicly available, it can be used to evaluate closed-source LLMs in the future, and we encourage others to do so.

\section{Acknowledgement}
This work was funded by the German Federal Ministry of Education and Research (BMBF) under the promotional reference 13N15897 (MISRIK), and the LOEWE initiative (Hesse, Germany) within the emergenCITY center. We would like to thank Sukannya Purkayastha, Aniket Pramanick, Ilia Kuznetsov, Kexin Wang, and Luke Bates for their constructive feedback and discussions of the work. We thank Sukannya Purkayastha, Jonathan Tonglet, and Yongxin Huang for their suggestions on a draft of the paper.

\bibliography{anthology}

\vspace{5pt}
\appendix
\section{Dataset}\label{app:dataset}

\subsection{Annotations}\label{app:examples}

\begin{figure}
    \centering
    \includegraphics[width=0.4\textwidth]{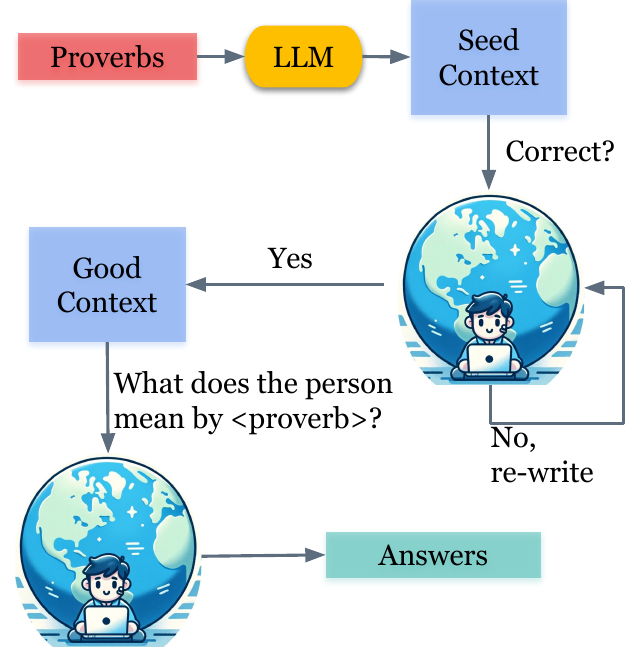}
    \caption{The data annotation process of \dataname{}.}
    \label{fig:annotation}
\end{figure}

We recruit crowd annotators through Prolific\footnote{\url{http://prolific.com/}} with the requirement of the corresponding language as their first language, and being fluent in English. Expert annotators are Master's, PhD and Post-doc researchers, including the authors of this paper. The annotation process is illustrated in Figure~\ref{fig:annotation}.

Instructions to create the conversational context:

\begin{mdframed}
\textbf{Step 1: Check if the proverb is used correctly in the conversation.}

Note: Sometimes, the proverb is figurative, meaning that the underlying meaning and the literal meaning of the proverb are different!
The conversation should fit the figurative usage/meaning of the proverb.
\\
\underline{Example}:

\textit{Person 1: "I'm scared of my boss." Person 2: "Well, barking dogs seldom bite."}

"Barking dogs seldom bite" -It has a literal meaning of dogs that bark rarely take action and bite you, so you don't need to be afraid of getting hurt.
The proverb metaphorically describes people that threaten you a lot rarely take action and harm you. Although this conversation may be missing some contexts, it should be labelled as correct.
\\
\underline{Example}:

\textit{Person 1: "My dog is barking." Person 2: "Well, barking dogs seldom bite."}

The proverb is used in a literal way when it has a figurative meaning.
This should be labelled as wrong.

\vspace{2pt}

\textbf{Step 2: Re-write the conversation if the proverb is not used correctly from step 1.}

The conversation should be 1-turn (1 round between 2 people), and maximum 2-turn (2 rounds between 2 people). 

Note: Please do not produce a conversation where one person is asking about the meaning of the proverb.

\end{mdframed}

\vspace{10pt}
Instructions to create the answers:

\begin{mdframed}
  What does the person mean?	
  \begin{itemize}
      \item Identify the person who used the proverb in the conversation.
      \item Write down a short sentence in the OPT1 column, state what the person means by the proverb in this conversation.
      \item Write down a negative of OPT1 in the OPT2 column.
  \end{itemize}
\end{mdframed}

\subsection{Animal and Food Terms in the Dataset}\label{app:data_analysis}

Table~\ref{tab:data_term} shows selected animal and food concepts across different languages. From the data, we can see that proverbs naturally contain culturally important concepts. For example, we can see that ``tiger'' is a relatively important concept for Eastern cultures, whereas ``lion'' is more important for Western cultures; while bread is enjoyed by many people around the world, rice is culturally more important in the East.

\begin{table}[t!]
    \centering
    \resizebox{\linewidth}{!}{%
    \begin{tabular}{ll}
    \includegraphics[width=\textwidth]{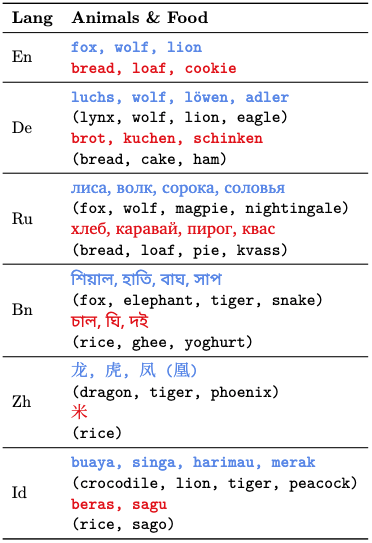}\\
    \end{tabular}}
    \caption{Selected \textcolor{nice-red}{food} and \textcolor{nice-blue}{animal} concepts from the proverbs.}
    \label{tab:data_term}
\end{table}

\subsection{Additional Qualitative Analysis of Proverbs}\label{app:qual_analysis}

We provide a qualitative analysis of how similar proverbs are expressed differently across languages and cultures. Similar to the ones in our introduction, many proverbs have a similar variant across cultures but are expressed differently. These proverbs differ by either using concepts that are familiar with the culture or using a local place name or person name (but this is very rare). Table~\ref{tab:sim_proverbs} shows examples. 

\begin{table}[t!]
    \centering
    \resizebox{\linewidth}{!}{
    \begin{tabular}{ll}
        \includegraphics[width=0.6\textwidth]{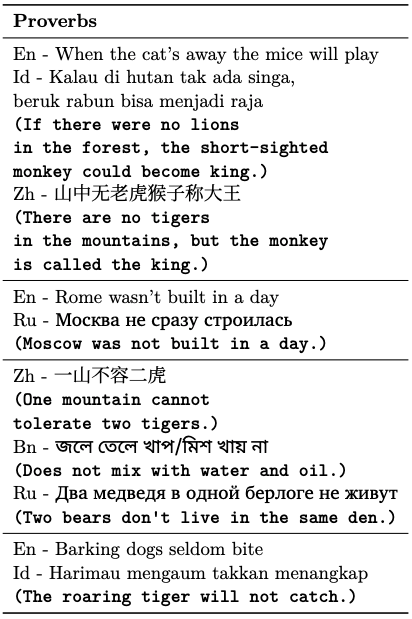}\\
    \end{tabular}
    }
    \caption{Parallel or closely related proverbs across different languages.}
    \label{tab:sim_proverbs}
\end{table}

\begin{figure*}[t!]
    \centering
    \includegraphics[width=0.9\textwidth]{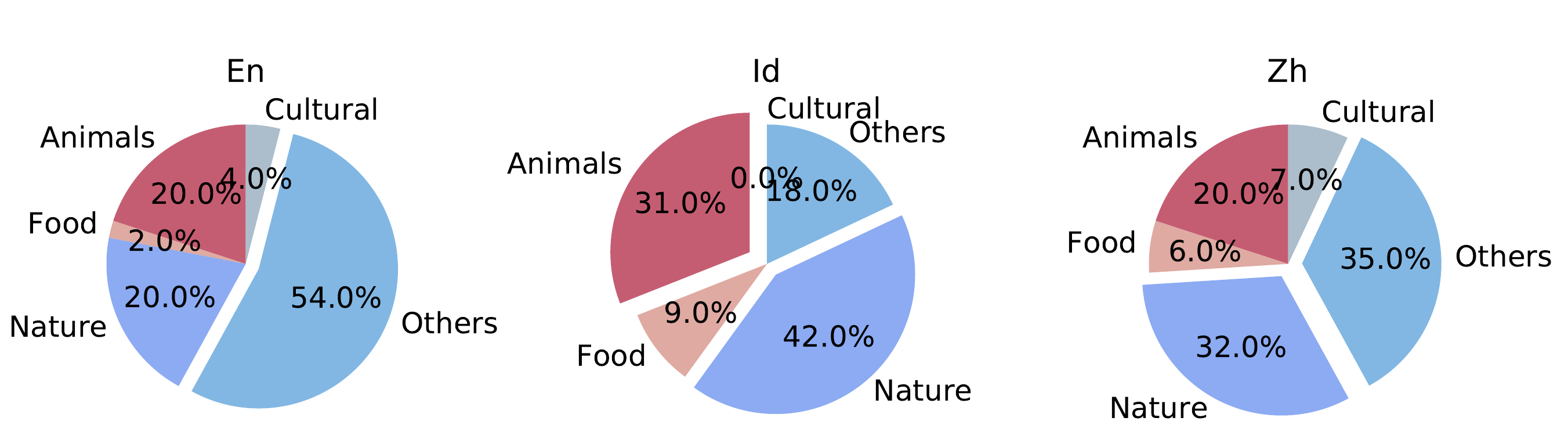}
    \caption{Distributions of concepts categories in figurative proverbs.}
    \label{fig:pie_fig}
\end{figure*}

Next, when proverbs are figurative, different languages and cultures tend to use different types of concepts to draw parallels. We randomly sampled 100 figurative proverbs in English, Indonesian and Chinese, and classified contained concepts into one of the 5 categories, namely: Animals \& Insects, Food, Cultural (including religious and spiritual entities, historical figures or names from the local culture), Nature (including metals, plants and other in-animated objects) and Others. Most of the time, a proverb only contains a single type of concept. However, when there are multiple types of concepts, we pick the dominant one (such as part of the object of the sentence). The distributions are in Figure~\ref{fig:pie_fig}. Here, we observe noticeable differences in distributions across different cultures. There are more concepts related to Animals \& Insects and Nature in Indonesian than in other languages, which is probably due to Indonesia's unique geographical location. 

\subsection{Additional Data Statistics}

We include additional dataset statistics in Table~\ref{tab:data_stats}. To calculate the average tokens in the context for Chinese, we take each character as a word. 

\begin{table}[t!]
    \centering
    \resizebox{0.9\linewidth}{!}{%
    \begin{tabular}{lcc}
    \toprule
        \textbf{Lang} & Avg Tok in Context & Avg Turns  \\
         \midrule
         English & 28.41 & 1.18\\
         Chinese  & 31.30 & 1.14\\
         German   & 27.91 & 1.12\\
         Indonesian & 25.35 & 1.15 \\
         Russian  & 31.25 & 1.47\\
         Bengali  & 35.16 & 1.63\\
    \bottomrule
    \end{tabular}
    }
    \caption{Additional dataset statistics: average number of tokens in the context, and average turns in the context.}
    \label{tab:data_stats}
\end{table}

\subsection{Interpreting the KDE Plot}\label{app:kde}

\begin{figure}

    \centering
    \includegraphics[width=0.48\textwidth]{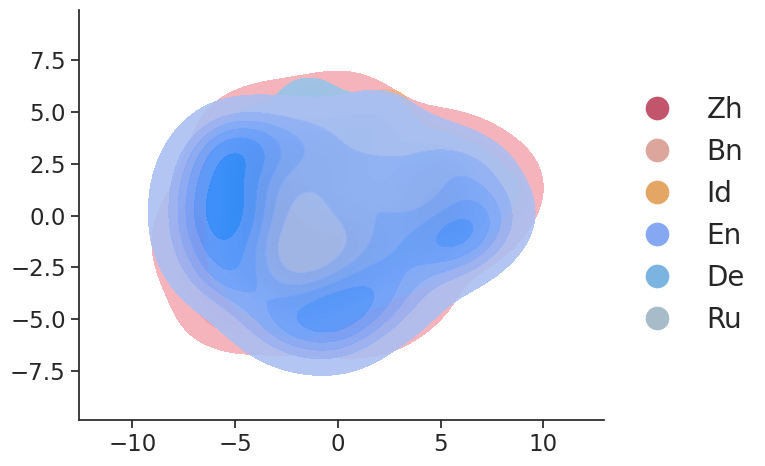}
    \caption{Visualizing embeddings with
Kernel Density Estimate (KDE) when the sentences are sampled from a parallel dataset (topic coherent across languages).}
    \label{fig:data_analysis_rand}
\end{figure}

For better comparison, we produce the Kernel Density
Estimate (KDE) plot of 400 randomly sampled sentences in each language (2400 sentences in total), from a parallel multilingual dataset~\cite{bactrianx} in Figure~\ref{fig:data_analysis_rand}. As the original data is much larger (67k sentences per language), sub-sampled sentences are likely not translations of each other, but rather topic-coherent. 

When sentences are topic-coherent, their embeddings overlap on top of each other and are inseparable (Figure~\ref{fig:data_analysis_rand}). In comparison with the KDE plot of proverb embeddings (Figure~\ref{fig:data_analysis}), we can see the difference in proverbs across languages and cultures.  

\subsection{Data Examples}\label{app:examples}

We balance the labels in \dataname{} and we show example data for all languages in Table~\ref{tab:all_examples}. 

\begin{table*}[t!]
    \centering
    \resizebox{\textwidth}{!}{%
    \begin{tabular}{c}
        \includegraphics[width=\textwidth]{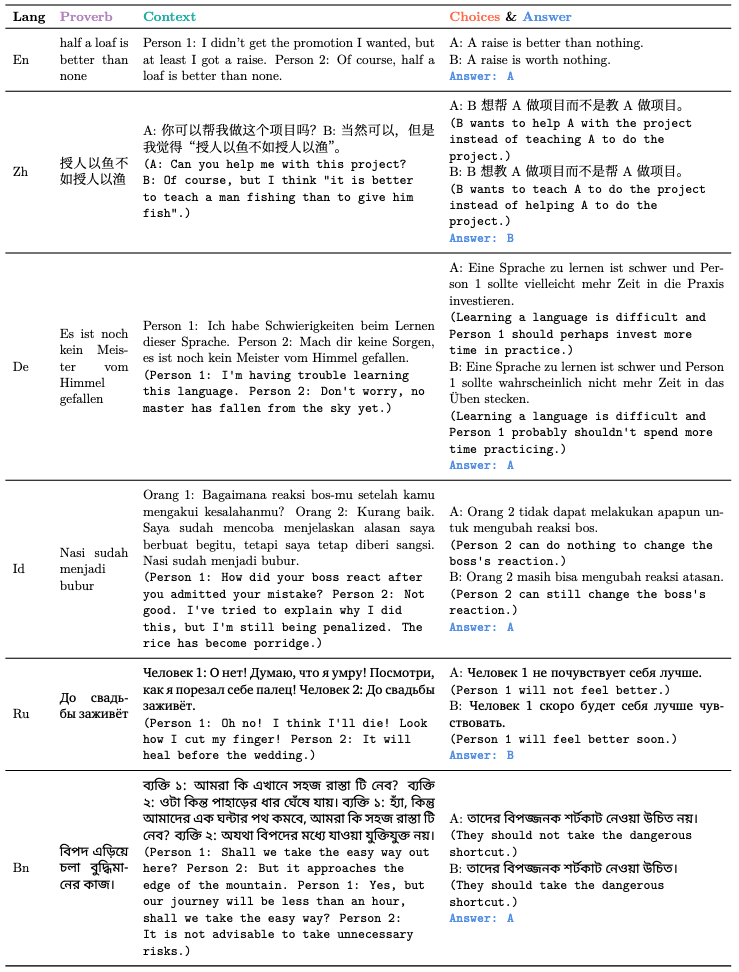}\\
    \end{tabular}}
    \caption{Examples for all six languages from \dataname{}.}
    \label{tab:all_examples}
\end{table*}

\section{Templates}\label{app:template}

We use \ex{Generate a very short 1-turn dialogue ends with ``{proverb}'' in {language}} as the template to query GPT3.5 (gpt-3.5-turbo-0301) for the seed conversational data. The model does not strictly generate seed conversation with 1-turn. We also experimented with a translated template and did not observe quality improvements for our task.  

Table~\ref{tab:prompt_temp_mem} contains all of the templates used in our memorization experiments in the main section, with the proverb \ex{no pain, no gain} as the example. For instance, the last word of \ex{no pain, no gain} is removed. As the prompting results are highly variable based on the input patterns, we created five different prompt patterns. We take the union of memorized examples among 5 patterns as the memorization accuracy.

\begin{table}
    \centering
    \resizebox{\linewidth}{!}{
    \begin{tabular}{l}
        \toprule
        \textbf{Templates} \\
        \midrule
         1. \textcolor{dblue}{\textbf{Proverb:}} \texttt{no pain, no } \\
         2. \textcolor{dblue}{\textbf{Complete this proverb:}} \texttt{no pain, no } \\
         3. \textcolor{dblue}{\textbf{Finish the proverb:}} \texttt{no pain, no } \\          
         4. \textcolor{dblue}{\textbf{What's the last word of this proverb:}} \texttt{no pain, no } \\
         5. \textcolor{dblue}{\textbf{What's missing at the end of this proverb:}}\\ \texttt{no pain, no } \\
       \bottomrule
    \end{tabular}
    }
    \caption{Memorization templates and the coloured portion is the template. }
    \label{tab:prompt_temp_mem}
\end{table}

\begin{figure}
    \centering
    \includegraphics[width=0.98\linewidth]{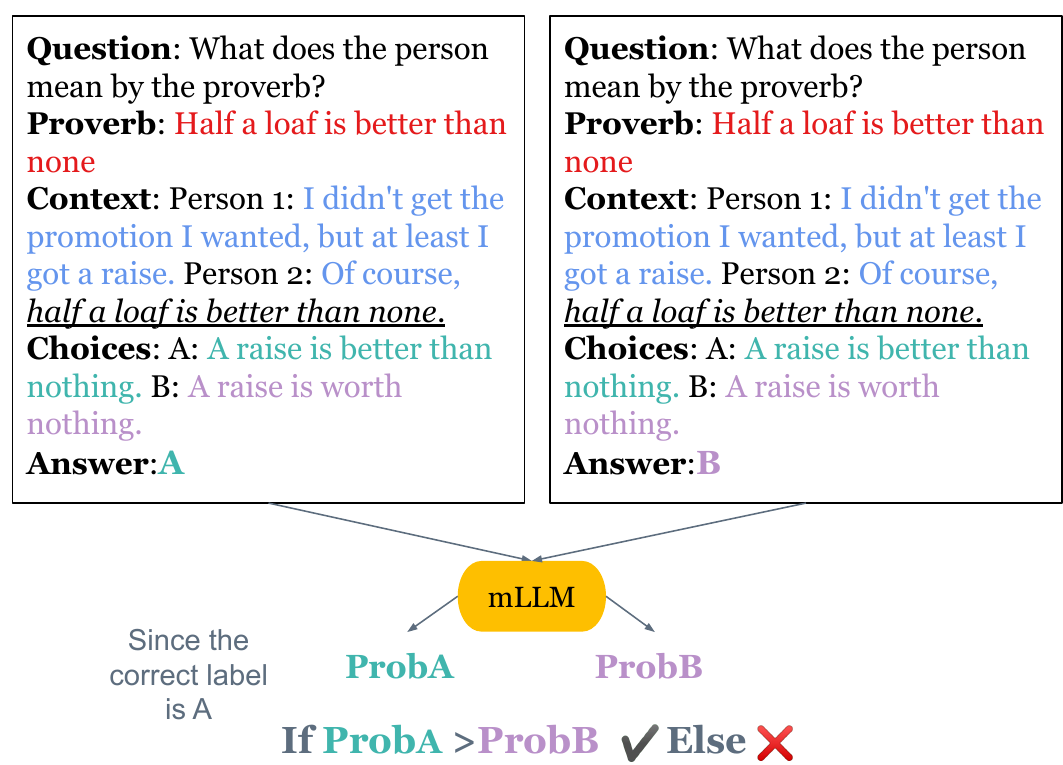}
    \caption{An example illustrating how the inference is done with mLLMs (excluding MLMs).}
    \label{fig:exp}
\end{figure}

Table~\ref{tab:prompt_temp} is the template we used for our main inference experiment in the paper. As described in \S\ref{sec:exp}, we perform experiments with the inference task on mLLMs. Figure~\ref{fig:exp} illustrates the experiment process for non-MLM models.

\begin{table}[t!]
    \centering
    \resizebox{0.88\linewidth}{!}{
    \begin{tabular}{l}
        \toprule
        \makecell*[{{p{7.8cm}}}]{
        \textcolor{dblue}{\textbf{Question:} What does the person mean by the proverb?}\\
        \textcolor{dblue}{\textbf{Proverb:}} <proverb> \\
        \textcolor{dblue}{\textbf{Context:}} <context>\\
        \textcolor{dblue}{\textbf{Choices:}} A: <answer 1> B: <answer 2> \\
        \textcolor{dblue}{\textbf{Answer:}}
        }\\\bottomrule
    \end{tabular}
    }
    \caption{Zero-shot testing template, where the coloured portion is the template.}
    \label{tab:prompt_temp}
\end{table}

\section{Cross-lingual Transfer Baselines}\label{app:xling_trans}

For completeness, we provide cross-lingual transfer baselines on \dataname{}. For cross-lingual transfer baselines, we re-split the English dataset into the train and test set (274/150 data points each) and evaluate on the original test set for other languages (i.e., same as zero-shot). We randomly sampled 20 data points from the training set as validation.  
We formulate the task as binary classification and experimented with XLM-R-Base (125M)/XLM-R-Large (355M)/XLM-R-XL (3.5B) and mT0-Base (580M)/mT0-Large (1.2B)/mT0-XL (3.7B). 

The data input format is: \ex{Context: \{context\} Choices: A: \{answer 1\} B: \{answer 2\}}. 

We use AdamW optimizer~\cite{adamw} and conduct a hyperparameter search of the learning rate of [5e-5, 1e-4, 1e-5] and batch size of [8, 10, 16], trained for 30 epochs with bfloat16 precision, on a single A100 GPU.

The zero-shot transfer results are in Table~\ref{tab:baselines} and averaged over 4 random seeds. The final hyperparameters for all models are [lr=1e-4, batch size=10], except for mT0-Large, which is [lr=1e-4, batch size=8]. Following previous work, we also include results for the translate-test baselines~\cite{conneau-etal-2018-xnli} in Table~\ref{tab:baselines}.

Similar to our findings in the main paper, the model does not perform well on the task with models under a billion parameters. The performance gap between English and other languages remains significant. 

\begin{table*}[]
    \centering
        \resizebox{0.9\linewidth}{!}{
        \tiny
    \begin{tabular}{l c cccccc}
    \toprule
    \textbf{Model} & \textbf{En} & \textbf{De} & \textbf{Zh} & \textbf{Ru} & \textbf{Id} & \textbf{Bn} & \textbf{Cross-lingual Avg}\\
    \midrule
         XLM-R-Base (125M) & 52.06  & 50.00 & 50.07 & 50.19 & 50.37 & 50.22 & 50.17 \\
         XLM-R-Large (355M) &  49.85 & 50.00 & 50.07 & 50.00 & 49.93 & 50.00 & 50.00\\
         XLM-R-XL (3.5B) &  58.38 &53.67& 52.25 & 53.65 & 52.79 & 53.01 & 53.07 \\
         \midrule
         mT0-Base (580M) & 60.74 & 55.01 & 52.02 & 50.77 & 50.29 & 53.75 & 52.37\\
         mT0-Large (1.2B) & 65.00 & 56.89 &  56.59 &  53.53 & 50.44 & 55.59 & 54.61\\
         mT0-XL (3.7B) & 72.65 & 67.51 & 60.63 & 61.54 & 60.26 & 53.82 & 60.75 \\
    \midrule
    \textit{Translate-Test} \\
    \midrule
         XLM-R-Base (125M) & - & 50.60 & 50.75 & 49.23 & 51.47 & 49.85 & 50.38\\
         XLM-R-Large (355M) & - & 50.00  &  50.00 & 50.00  &  49.85  &  50.00  & 49.97 \\
         XLM-R-XL (3.5B) & - & 50.90	& 51.20	& 52.31	& 49.85	& 51.47	& 51.15 \\
         \midrule
         mT0-Base (580M) & - & 51.80 & 51.05 & 51.15 & 49.56 & 54.26 & 51.56\\
         mT0-Large (1.2B) & - & 54.04 & 55.09 & 54.62 & 53.67 & 57.21 & 54.93\\
         mT0-XL (3.7B) & - & 67.96 & 62.72 & 63.46 & 57.92 & 58.68 & 62.15 \\
         
    \bottomrule
    \end{tabular}
    }
    \caption{Zero-shot cross-lingual transfer and translate-test baselines. Cross-lingual averages are calculated over all languages except English.}
    \label{tab:baselines}
\end{table*}

\section{Additional Results}

\subsection{Further Details on the Memorization Experiment}\label{app:add_mem}

\subsubsection{Discussions}\label{app:add_mem_disc}

Following the defined criteria for identifying memory recall from LLMs in \cite{haviv-etal-2023-understanding}, a generalized prediction by LM always has alternatives based on the context to express similar meanings. Specifically, a phrase like \ex{no pain, no \textunderscore\textunderscore} would elicit multiple possible predictions. Without knowing the proverb, words like ``painkiller", ``medication" or ``suffrage" are highly likely to occur at the last position based on the context. Similarly, a phrase such as \ex{It is better to teach a man fishing than to give him \textunderscore\textunderscore}, similar concepts like ``food", ``Carps", or even ``money" are very reasonable.
Hence, a LLM that predicts the correct missing word (the single correct prediction) has likely memorized the data.


Certainly, based on the training method of LLMs, an alternative setup could be to mask words at various locations and have the model predict the missing words. However, such a method is more suitable to mT0 and XLM-R, which explicitly incorporate masked token predictions in pretraining with \texttt{<extra\_id\_0>}~\cite{DBLP:journals/jmlr/RaffelSRLNMZLL20} tokens or \texttt{<mask>} tokens.

\subsubsection{Additional Results}\label{app:add_mem_res}

\begin{figure}
    \centering
    \includegraphics[width=0.98\linewidth]{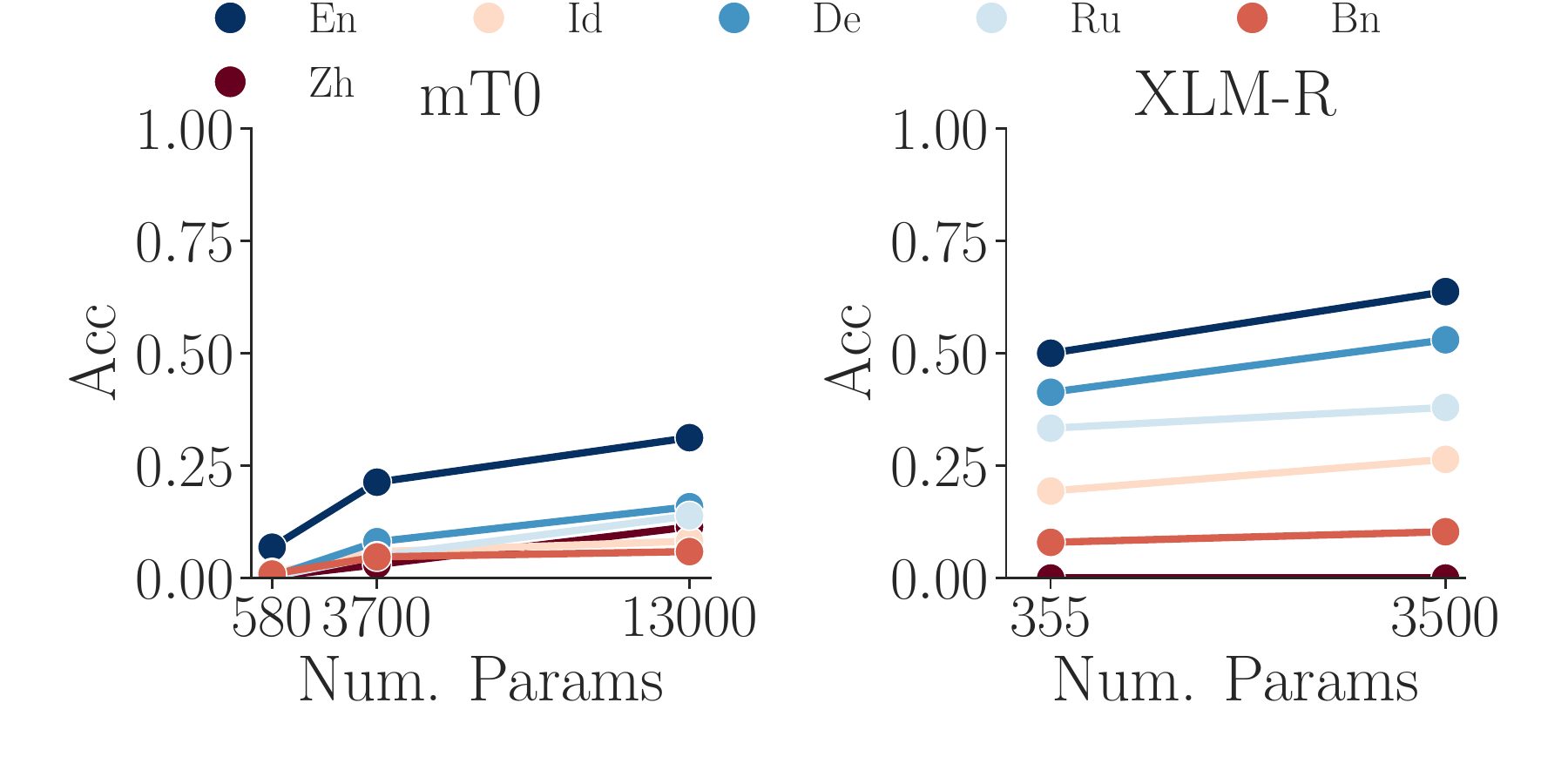}
    \caption{Memorization of proverbs in different languages when masking out words randomly.}
    \label{fig:prompt_temp_mem2}
\end{figure}

We include the results with randomly masked tokens (1 masked token per proverb) here for completeness for mT0 and XLM-R models. However, we use modified prompts in Table~\ref{tab:prompt_temp_mem2} due to our prior prompts were constructed for predicting the last word of the proverbs, and with \texttt{<extra\_id\_0>} or  \texttt{<mask>} as the masked token for mT0 and XLM-R respectively. The results are in Figure~\ref{fig:prompt_temp_mem2}. Several observations still persist, such as the disparity in memorization between languages, the low memorization rate of mT0 models, and the positive correlation between model size and memorization, etc.

\begin{table}
    \centering
    \resizebox{\linewidth}{!}{
    \begin{tabular}{l}
        \toprule
        \textbf{Templates} \\
        \midrule
        1. \textcolor{dblue}{\textbf{Fill the missing token:}} \texttt{no <mask>, no gain}  \textcolor{dblue}{\textbf{Answer:}} \\
        2. \textcolor{dblue}{\textbf{What is the missing word in this proverb:}} \texttt{no <mask>, no gain}  \textcolor{dblue}{\textbf{Answer:}} \\
        3. \textcolor{dblue}{\textbf{What is the masked word in this proverb:}} \texttt{no <mask>, no gain}  \textcolor{dblue}{\textbf{Answer:}} \\
       \bottomrule
    \end{tabular}
    }
    \caption{Additional memorization templates, adjusted the task description to fit the experiment with random words removed from the proverb. The coloured portion is the template. \texttt{<extra\_id\_0>} is used for mT0 and  \texttt{<mask>} is used for XLM-R.}
    \label{tab:prompt_temp_mem2}
\end{table}

\subsection{Memorized versus Not Memorized}\label{app:mem}

We break down the results into memorized groups versus not memorized groups for the three best-performing models. We only show results when there are more than 50 proverbs in a group in Table~\ref{tab:mem_vs_reason} (which left us with English and Chinese). The benefit of memorization only shows for English, but not for Chinese. 

\begin{table}[]
    \centering
        \resizebox{\linewidth}{!}{

    \begin{tabular}{l|cccc}
    \toprule
    & \multicolumn{2}{c}{\textbf{En}}  & \multicolumn{2}{c}{\textbf{Zh}}  \\
    \textbf{Model}  & $\in$Mem. &$\notin$Mem. & $\in$Mem. &$\notin$Mem. \\
    \midrule
        BLOOMZ 7.1B & 77.23 & 65.07 & - & - \\
        mT0-XXL (13B)     & 86.17 & 84.33 & 81.48 & 82.50 \\
        LLaMA-2 13B     & 80.30 & 75.38 & 54.65 & 53.22\\
    \bottomrule
    \end{tabular}
    }
    \caption{Result on memorized versus not memorized proverbs on 3 best performing models for English and Chinese. Results were omitted due to less than 50 proverbs in the not memorized group.}
    \label{tab:mem_vs_reason}
\end{table}

\subsection{Additional Results for the Inference Task}

\subsubsection{Additional Prompt Templates}\label{app:add_pos}
\begin{figure*}
    \centering
    
    \begin{subfigure}[b]{\linewidth}
         \centering
         \includegraphics[width=\linewidth]{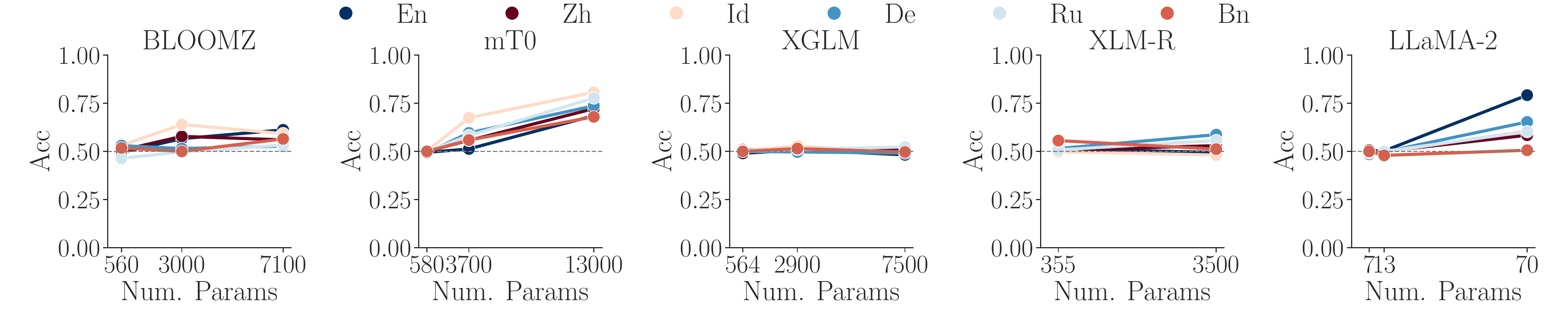}
         \caption{Additional results using \ex{Temp1} in Table~\ref{tab:prompt_temp2}.}
         \label{fig:addtemp1}
     \end{subfigure}
     
    \hfill
    \vspace{0.2cm}
    \begin{subfigure}[b]{\linewidth}
         \centering
         \includegraphics[width=\linewidth]{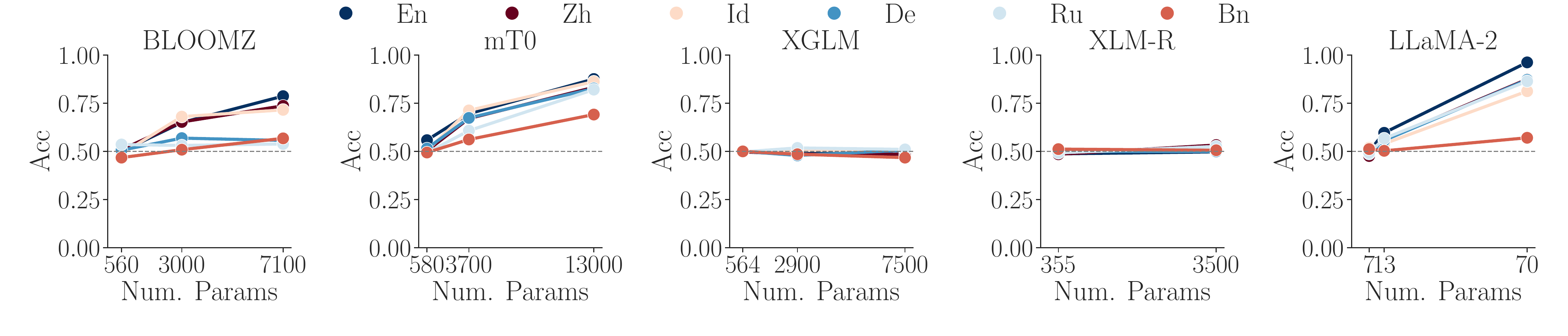}
         \caption{Additional results using \ex{Temp2} in Table~\ref{tab:prompt_temp2}.}
         \label{fig:addtemp2}
     \end{subfigure}

    \hfill
    \vspace{0.2cm}
    \begin{subfigure}[b]{\linewidth}
         \centering
         \includegraphics[width=\linewidth]{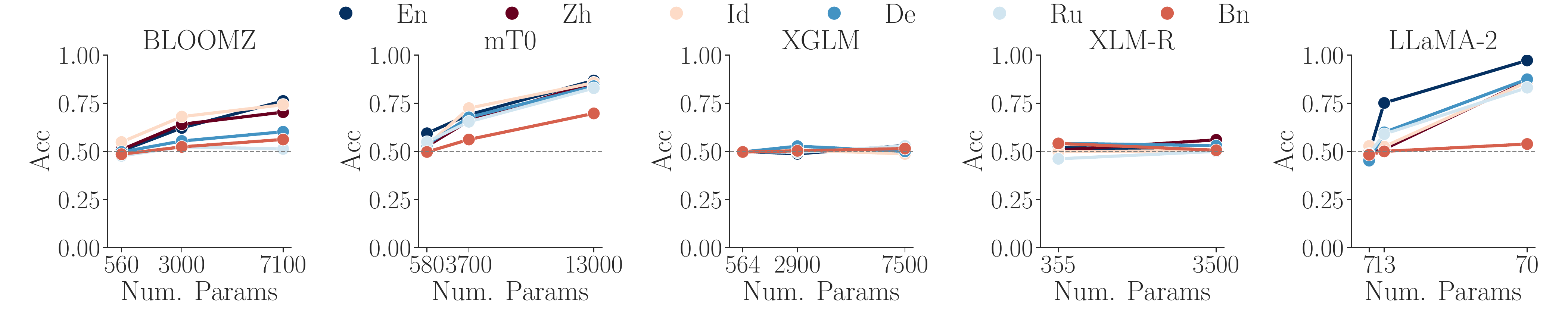}
         \caption{Additional results using \ex{Temp3} in Table~\ref{tab:prompt_temp2}.}
         \label{fig:addtemp3}
     \end{subfigure}

    \caption{Performance of mLLMs on the proposed \dataname{} dataset with additional templates. The number of parameters is in billions for
LLaMA-2 and in millions for all other models.}
    \label{fig:add_temp}
\end{figure*}

We experimented with 3 additional prompt templates in Table~\ref{tab:prompt_temp2} to demonstrate the generality of our findings. Our experiments (Figure~\ref{fig:add_temp}) show similar trends as in Figure~\ref{fig:qa_pos} of the main section of our paper. We continue to observe that mT0 models perform the best for the inference task, the results improve as the model size increases, and memorization of the proverb is not an indication of performance on the inference task. However, we'd like to point out that our experiments do not assess the formal reasoning abilities~\cite[such as mathematical reasoning etc.]{huang-chang-2023-towards} of mLLMs.

\begin{table}
    \centering
    \resizebox{\linewidth}{!}{
    \begin{tabular}{l}
        \toprule
        \textbf{Additional Templates} \\
        \midrule
        \makecell*[{{p{7.8cm}}}]{
        Temp1:\\
        \textcolor{dblue}{\textbf{Proverb:}} <proverb> \\
        \textcolor{dblue}{\textbf{Context:}} <context>\\
        \textcolor{dblue}{\textbf{Choices:}} A: <answer 1> B: <answer 2> \\
        \textcolor{dblue}{\textbf{Question:} What does the person mean by the proverb?}\\
        \textcolor{dblue}{\textbf{Answer:}}
        }\\
        
        \midrule
        \makecell*[{{p{7.8cm}}}]{
        Temp2:\\
        \textcolor{dblue}{\textbf{Question:} What is a probable interpretation of this proverb?}\\
        \textcolor{dblue}{\textbf{Proverb:}} <proverb> \\
        \textcolor{dblue}{\textbf{Context:}} <context>\\
        \textcolor{dblue}{\textbf{Choices:}} A: <answer 1> B: <answer 2> \\
        \textcolor{dblue}{Please choose between A and B.} \\
        \textcolor{dblue}{\textbf{Answer:}}
        }\\
        
        \midrule
        \makecell*[{{p{7.8cm}}}]{
        Temp3:\\
        \textcolor{dblue}{\textbf{Question:} How would one interpret this proverb given the context?}\\
        \textcolor{dblue}{\textbf{Proverb:}} <proverb> \\
        \textcolor{dblue}{\textbf{Context:}} <context>\\
        \textcolor{dblue}{\textbf{Choices:}} A: <answer 1> B: <answer 2> \\
        \textcolor{dblue}{\textbf{Answer:}}
        }\\
       \bottomrule
    \end{tabular}
    }
    \caption{Additional prompt templates. The coloured portion is the template.}
    \label{tab:prompt_temp2}
\end{table}

\subsubsection{Additional Models}\label{app:add_models}

We include results of the additional following models, including Vicuna-V1.5~\cite[7B, 13B]{vicuna1p5}, and Aya-101~\cite[13B]{ayamodel}. 

Similar to what we observe in the main paper, as the model increases in size, the performance on \dataname{} is better when asking the model to pick the correct answer. Aya-101 model's performance is on par with mT0 13B in most of the languages (both when asking positive and negative questions), but Aya-101 is noticeably better in Bengali. 

\begin{figure}
    \centering
    \begin{subfigure}[b]{\linewidth}
         \centering
         \includegraphics[width=0.98\linewidth]{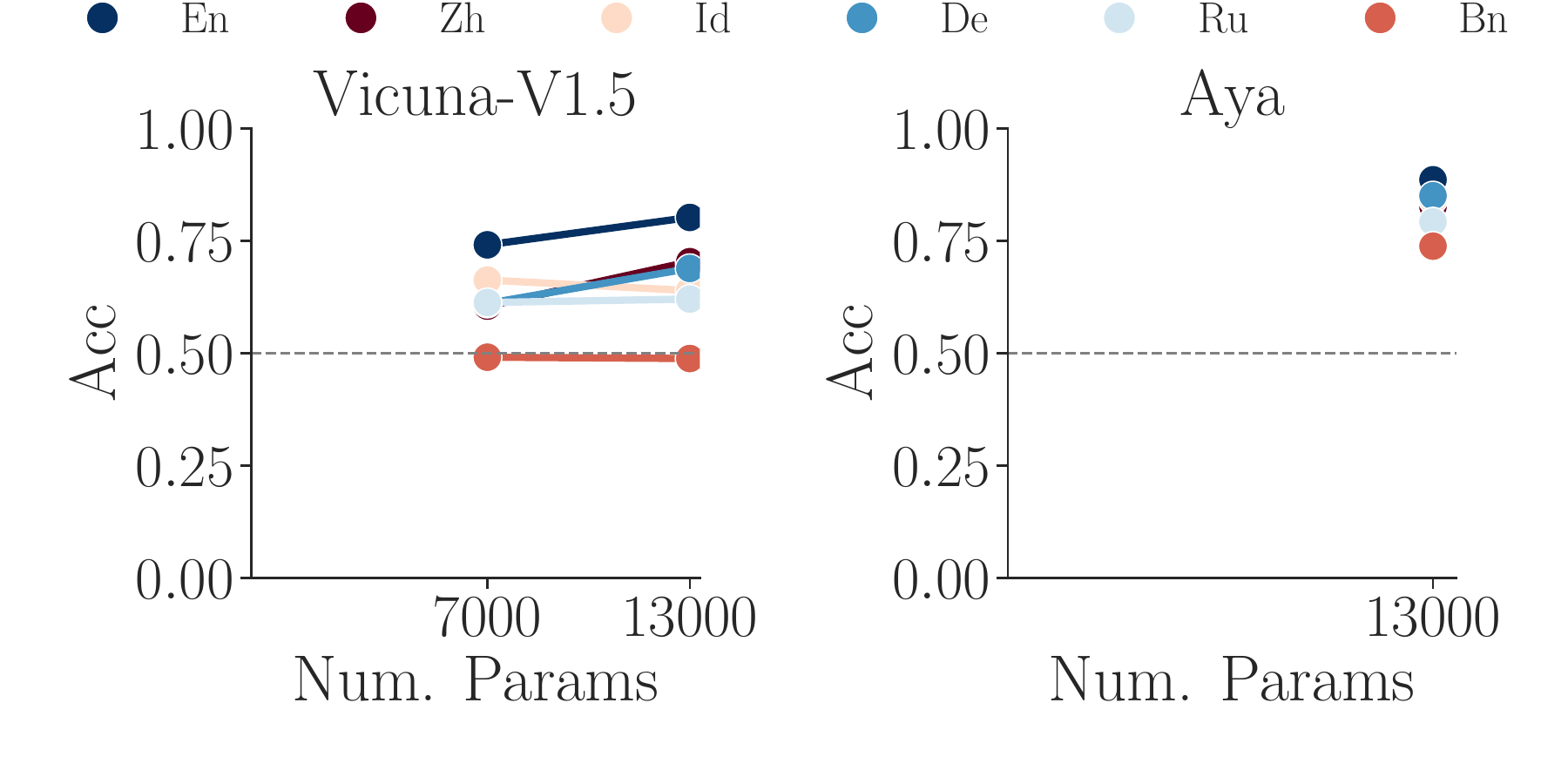}
         \caption{Performance of mLLMs on the proposed \dataname{}.}
         \label{fig:neg1}
     \end{subfigure}
    \hfill
    \vspace{0.2cm}
    \begin{subfigure}[b]{\linewidth}
         \centering
        \includegraphics[width=0.98\linewidth]{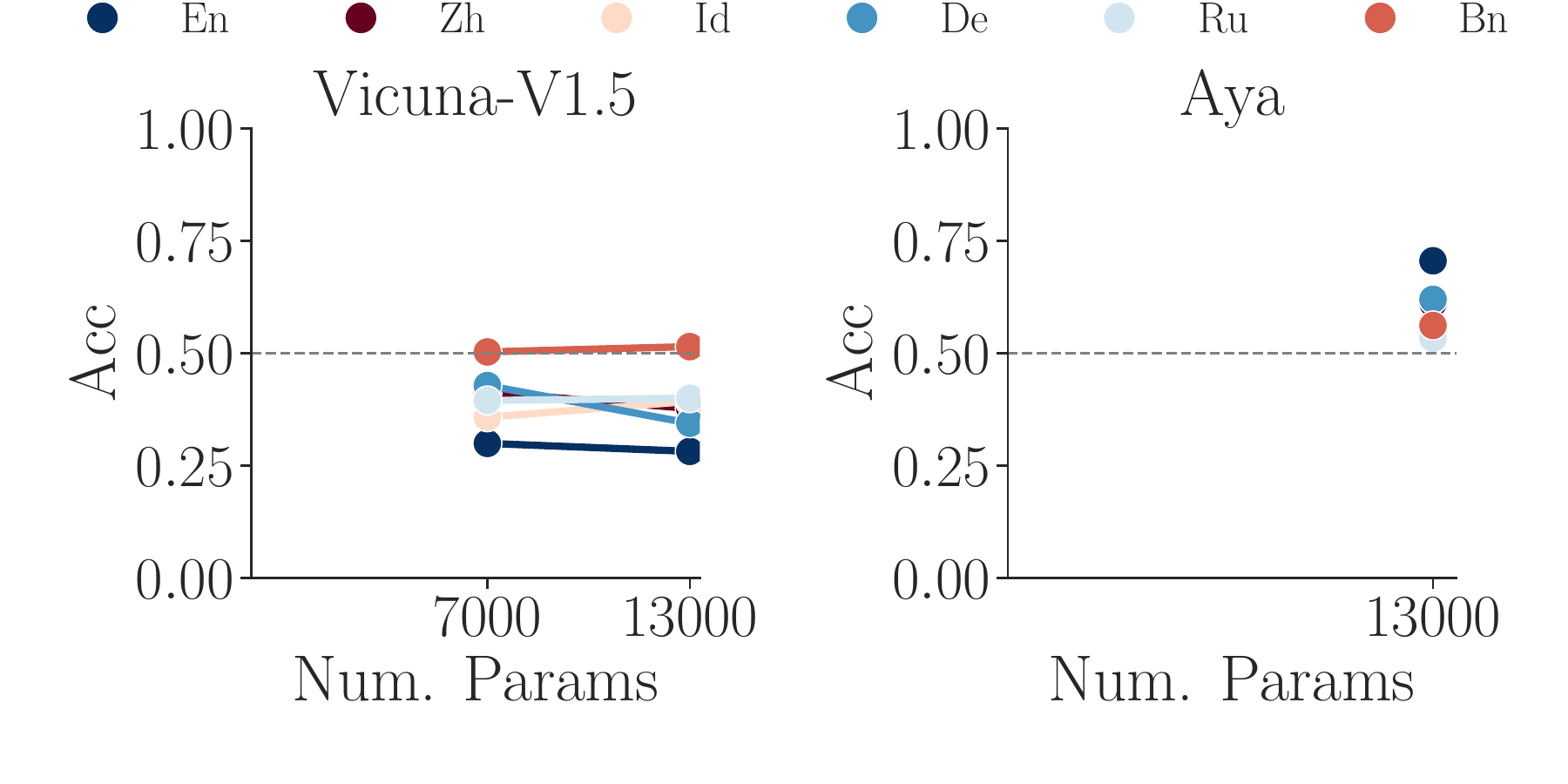}
        \caption{Performance of mLLMs on the proposed \dataname{} - Inference task when asking the `negative' question.}
         \label{fig:neg2}
     \end{subfigure}

    \caption{Performance of additional mLLMs on the proposed \dataname{} dataset. The number of parameters is in millions.}
    \label{fig:negs}
\end{figure}

\subsection{`\textit{Negative}' Questions}\label{app:negation}

We experimented with 4 additional versions of `\textit{negative}' questions/instructions (randomly created), without the use of the word \textbf{`not'}, they are:

\begin{itemize}
    \item Which answer is contrary to what the person means by the proverb?
    \item Which answer is impossible as the interpretation of what the person means by the proverb?
    \item Pick the opposite answer to what the person means by the proverb.
    \item Pick the wrong answer to what the person means by the proverb.
\end{itemize}

We use the same prompt template to evaluate the models. The results are in Figure~\ref{fig:negs}. While our work focuses on reasoning with cultural common grounds, this shows the importance and urgent need to improve the model's ability to answer `negative' questions. 

\begin{figure*}[t!]
    \centering
    \begin{subfigure}[b]{\linewidth}
         \centering
         \includegraphics[width=\linewidth]{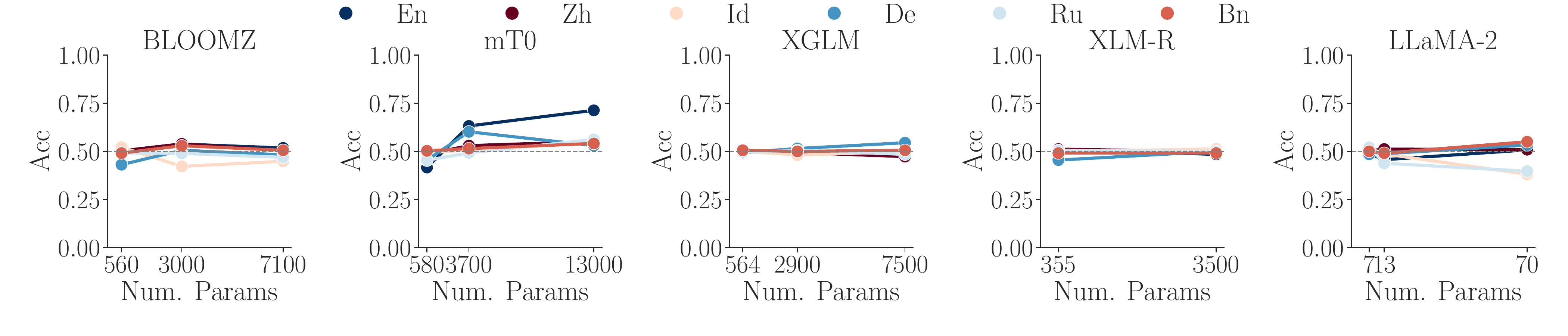}
         \caption{Results using \ex{Which answer is contrary to what the person means by the proverb?}.}
         \label{fig:neg1}
     \end{subfigure}
    \hfill
    \vspace{0.2cm}
    \begin{subfigure}[b]{\linewidth}
         \centering
        \includegraphics[width=\linewidth]{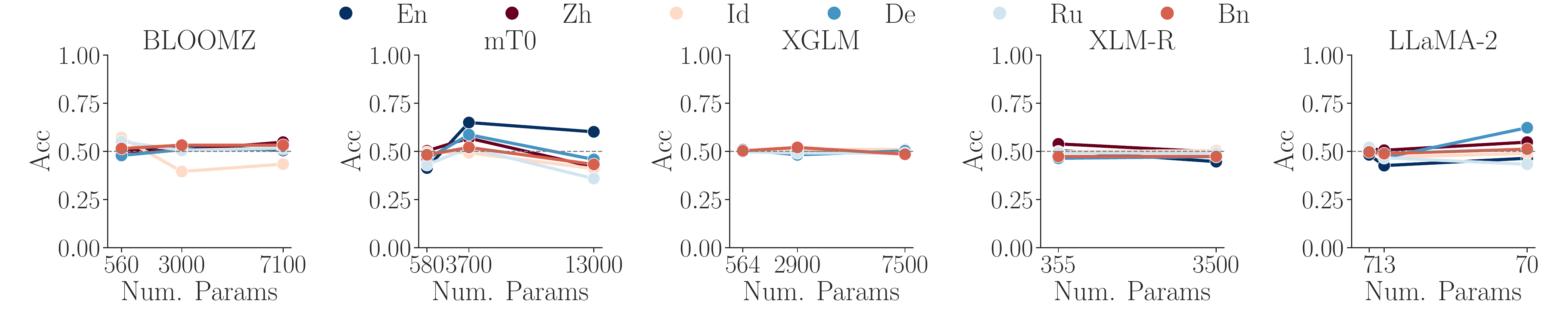}
        \caption{Results using \ex{Which answer is impossible as the interpretation of what the person means by the proverb?}.}
         \label{fig:neg2}
     \end{subfigure}

    \hfill
    \vspace{0.2cm}
    \begin{subfigure}[b]{\linewidth}
         \centering
        \includegraphics[width=\linewidth]{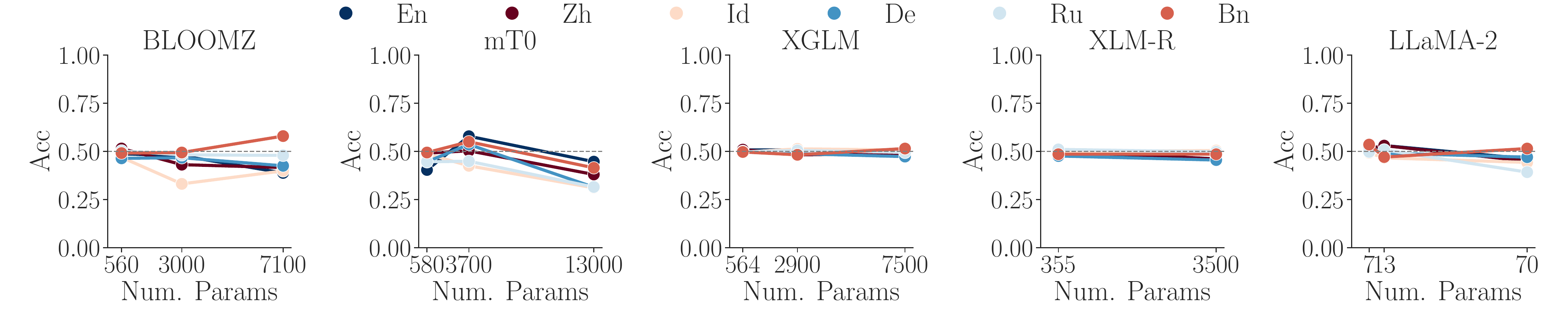}
        \caption{Results using \ex{Pick the opposite answer to what the person means by the proverb}.}
         \label{fig:neg3}
     \end{subfigure}

        \hfill
    \vspace{0.2cm}
    \begin{subfigure}[b]{\linewidth}
         \centering
        \includegraphics[width=\linewidth]{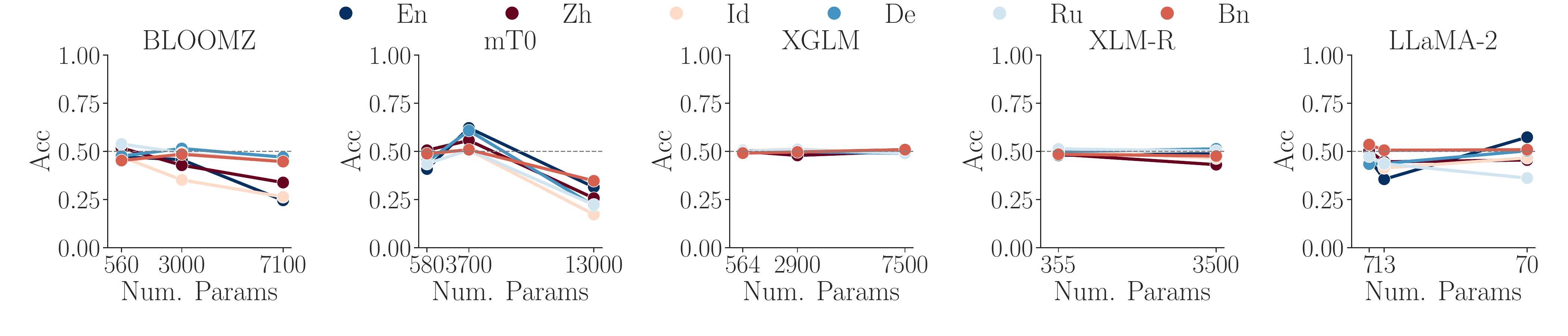}
        \caption{Results using \ex{Pick the wrong answer to what the person means by the proverb}.}
         \label{fig:neg4}
     \end{subfigure}

    \caption{Performance of mLLMs on the proposed \dataname{} dataset when asking the model a `negative' question. The number of parameters is in billions for
LLaMA-2 and in millions for all other models.}
    \label{fig:negs}
\end{figure*}

\begin{figure*}[t!]
    \centering
    \includegraphics[width=0.72\linewidth]{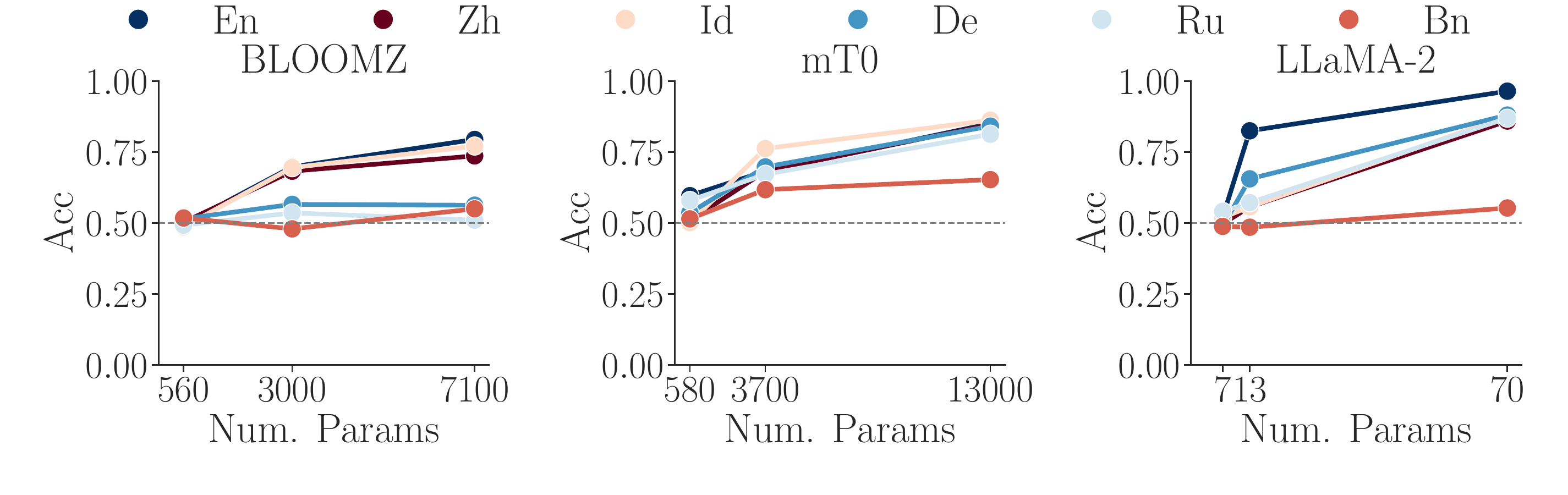}
    \caption{Performance of mLLMs on the proposed \dataname{} dataset when only the proverb, context and choices are provided, but without a question. Ideally, all models should score around random guessing. The number of parameters is in billions for
LLaMA-2 and in millions for all other models.}
    \label{fig:noq}
\end{figure*}

We speculate this is due to the biases in training data. Often, users seek the correct solution to solve problems online (which we refer to as positive biases) rather than the wrong solution. Hence, when using web corpora as training data for LLMs, such positive biases will propagate to the behaviour of LLMs. To demonstrate this further, we conducted an additional experiment \textit{without} asking a question in the prompt on BLOOMZ, mT0 and LLaMA-2. In an ideal situation, a good model should score nearly random when no question is asked (analogously to human confusion when data is given, but no question is asked). From Figure~\ref{fig:noq}, all LLMs can score above random for multiple languages, which indicates all models \textit{failed}. This failure mode further hints at the inability of mLLMs to handle negative questions maybe due to the nature of the training data. 

\subsection{Culture Gaps}\label{app:trans_test}

\begin{figure}[t!]
    \centering
    \includegraphics[width=0.5\textwidth]{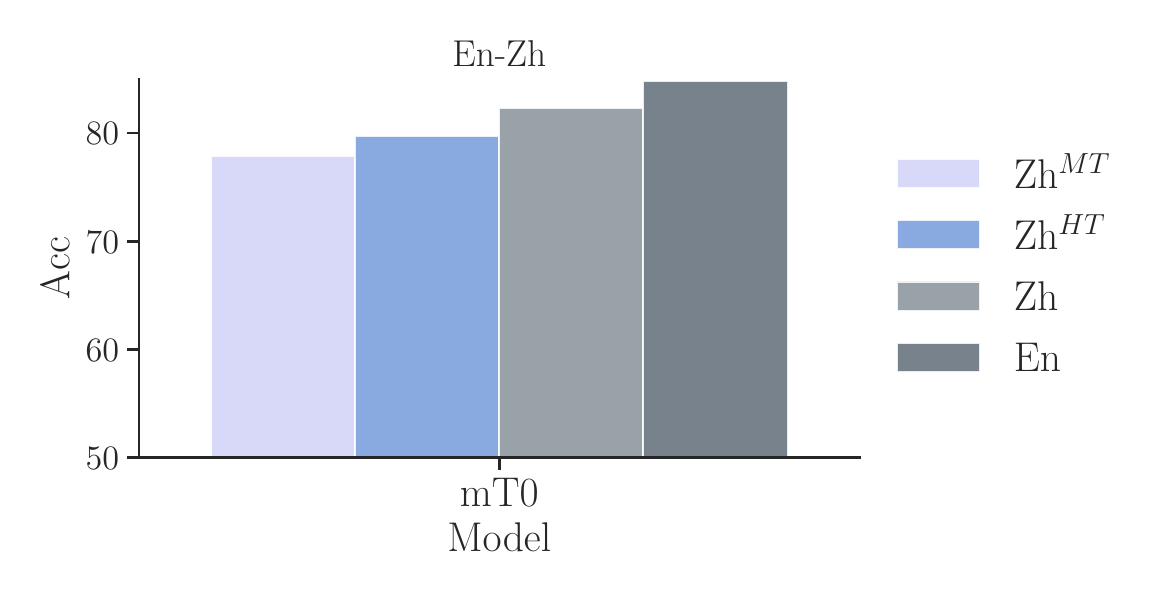}
    \caption{Performance gap between machine-translated, human-translated English data and results in the source language (En), and target language (Zh).}
    \label{fig:en2zh_cult_gap}
\end{figure}

In addition to the results in \S\ref{sec:cult_gap}, we follow the same procedure and perform the experiment with mT0 for En--Zh translated data. We observe similar results in Figure~\ref{fig:en2zh_cult_gap}, and the culture gap for En--Zh is 5.33.

\subsection{Additional Results on LLaMA-2 with Translations}\label{app:nshot}

Since LLaMA-2 13B is one of the recent state-of-the-art (English officially) models, we further conducted a zero-shot experiment by translating all data from other languages into English. We used Google Translate for translation and reported the results in Table~\ref{tab:trans_test}. From the Table, we can see significant performance gaps (to English). It is also interesting to see the gaps increase as the corresponding geographical location of the language moves further away from English. While we consider this gap to be a combination of the language gap and the defined culture gap, a future interesting direction is to closely examine the culture gap in cross-cultural communications and how this is related to the internal representations in LLMs.

\begin{table}[t!]
    \centering
        \resizebox{0.9\linewidth}{!}{
        \tiny
    \begin{tabular}{l cc c}
    \toprule
    \textbf{Lang} & \textbf{Ori. Lang} & \textbf{MT} & \textbf{$\Delta_{En}$} \\
    \midrule
    En & 78.68 & --- & --- \\
    De & 68.26 & 73.35 & 5.33\\
    Ru & 62.82 & 71.02 & 7.66 \\
    Id & 57.47 & 69.79 & 8.89 \\
    Bn & 49.11 & 61.76 & 16.92\\
    Zh & 53.59 & 54.19 & 24.49 \\
    \bottomrule
    \end{tabular}
    }
    \caption{Results of machine-translated data with LLaMA-2 13B. $\Delta_{En}$ is the resulting gap to the model's performance on English data.}
    \label{tab:trans_test}
\end{table}

\subsection{Few-shot (In-context) Evaluation}\label{app:nshot}

For completeness, we also provide evaluation results with few-shot demonstrations. We perform 2-shot and 5-shot experiments by randomly sampling 5 sets of n-shot demonstrations from the few-shot training set (using the same template as zero-shot evaluation by concatenation).  We evaluate BLOOMZ 7.1B, mT0-XXL 13B and LLaMA-2 13B models, and Table~\ref{tab:nshot} shows the results. 

From Table~\ref{tab:nshot}, we do not observe any improvements with few-shot demonstrations compared to zero-shot. In fact, model performances consistently degrade with more demonstrations. Since our task has a very long context that may affect the n-shot performance. Nonetheless, this degradation has been observed recently in other work such as in~\citet{cmmlu, indommlu} with few-shot evaluations. 

\begin{table*}[t!]
    \centering
        \resizebox{0.9\linewidth}{!}{
        \tiny
    \begin{tabular}{l c cccccc}
    \toprule
    \textbf{Model} & \textbf{En} & \textbf{De} & \textbf{Zh} & \textbf{Ru} & \textbf{Id} & \textbf{Bn} & \textbf{Cross-lingual Avg}\\
    \midrule
         BLOOMZ 7.1B 2-shot & 59.49 & 61.55 & 56.59 & 53.77 & 51.53 & 50.00 & 52.65 \\
         BLOOMZ 7.1B 5-shot & 51.57 & 52.39 & 50.85 & 50.35 & 50.25 & 50.52 & 50.30 \\
         \midrule
         mT0-XXL 13B 2-shot & 78.37 & 72.63 & 76.95 & 78.74 & 74.87 & 63.82 & 76.81 \\
         mT0-XXL 13B 5-shot & 68.48 & 67.90 & 70.38 & 71.50 & 67.64 & 60.00 & 69.57 \\
         \midrule
         LLaMA-2 13B 2-shot & 74.87 & 56.52 & 55.42 & 60.77 & 56.76 & 51.00 & 58.77 \\
         LLaMA-2 13B 5-shot & 64.16 & 52.69 & 54.89 & 55.56 & 52.71 & 50.17 & 54.14 \\
    \bottomrule
    \end{tabular}
    }
    \caption{Few-shot evaluation results from \dataname{}. Cross-lingual averages are calculated over all languages except English.}
    \label{tab:nshot}
\end{table*}

\end{document}